\documentclass[11pt]{article}

\usepackage[preprint]{acl}

\usepackage{times}
\usepackage{latexsym}

\usepackage[T1]{fontenc}
\usepackage[utf8]{inputenc}

\usepackage{microtype}
\usepackage{inconsolata}

\usepackage{graphicx}

\usepackage{amsfonts}
\usepackage{amsmath}
\usepackage{amssymb}
\usepackage{booktabs}
\usepackage{colortbl}
\usepackage{float}
\usepackage{url}
\usepackage{xspace}
\usepackage[ruled,vlined]{algorithm2e}
\usepackage{listings}

\usepackage{tikz}
\usetikzlibrary{arrows.meta}
\definecolor{scopeink}{HTML}{1F3A5F}
\newcommand{\scfull}{\tikz[baseline=-0.6ex]\fill[scopeink] (0,0) circle (0.36em);}
\newcommand{\schalf}{\tikz[baseline=-0.6ex]{\draw[scopeink,line width=0.5pt] (0,0) circle (0.36em);\fill[scopeink] (0,0) -- (90:0.36em) arc (90:270:0.36em) -- cycle;}}
\newcommand{\scnone}{\tikz[baseline=-0.6ex]\draw[scopeink!45,line width=0.5pt] (0,0) circle (0.36em);}
\definecolor{studyblue}{RGB}{233,241,248}
\definecolor{studyheader}{RGB}{213,226,237}
\definecolor{studystripe}{RGB}{246,248,250}
\newcommand{\na}{\textit{n/a}}
\newcommand{\sig}[1]{\rlap{$^{#1}$}\phantom{$^{**}$}}

\newcommand{\PAIR}{\textsc{Pair}}
\newcommand{\USER}{\textsc{User}}
\newcommand{\AGENT}{\textsc{Agent}}
\newcommand{\GLOBAL}{\textsc{Global}}
\newcommand{\bench}{\textsc{CrossLearnBench}}

\newcommand{\users}{\mathcal{U}}
\newcommand{\agents}{\mathcal{A}}

\newcommand{\method}{\textsc{MixUA}\xspace}

\makeatletter
\newcommand{\promptbox}[2]{\par\noindent\fcolorbox{black!35}{black!3}{\begin{minipage}{\dimexpr\linewidth-2\fboxsep-2\fboxrule\relax}\small\textbf{#1}\par\smallskip #2\end{minipage}}\par\medskip}
\newcommand{\pr}[1]{\textit{#1}\enspace}
\makeatother

\title{Who Remembers What? Multi-Granularity Memory Extraction for\\Multi-User, Multi-Agent Interaction}

\author{Yunzhe Wang$^{1,2}$\thanks{This work was done when Yunzhe Wang was an intern at Amazon. $^{**}$Project lead.}, Mesut Erhan Unal$^{1}$, Sheng Zhang$^{1}$, Shabnam Ghadar$^{1}$, Peng Tang$^{1**}$ \\
  $^{1}$AWS Agentic AI \quad $^{2}$University of Southern California \\
  \texttt{\{yunzhewa,meunal,szhangml,shabnam,tangpen\}@amazon.com}}

\begin{document}

\maketitle

\begin{abstract}
  Self-evolving agents improve their own skills, memory, and harness code from experience without retraining the base LLM, and group-evolving agents go further by sharing these lessons across agents.
  Yet most such systems overlook that sharing across users and heterogeneous agents requires deciding who can benefit from each lesson and what information must remain private.
  We present \method{}, a memory system that distills independent user-agent trajectories into lessons and assigns each lesson to a user-agent pair, user, agent, or global scope according to whom it benefits and what it may disclose.
  For evaluation, we introduce \bench{}, a mixed-user-agent benchmark that pairs established agent tasks with user preferences and privacy probes that test whether private facts leak into memory other users can read.
  With five users and seven agents, \method{} raises pass rates by up to 16 points over no memory and 5 over a single global memory, even on tasks previously tried only by other users on other agents. It retains up to 90\% of previously stated preferences, and at most 0.02\% of privacy probes find a planted private fact in a forbidden scope.
  We hope \method{} offers a scalable path toward a global system where users and agents co-evolve.
\end{abstract}

\section{Introduction}
\label{sec:intro}

\begin{figure}[t]
  \centering
  \includegraphics[width=\linewidth]{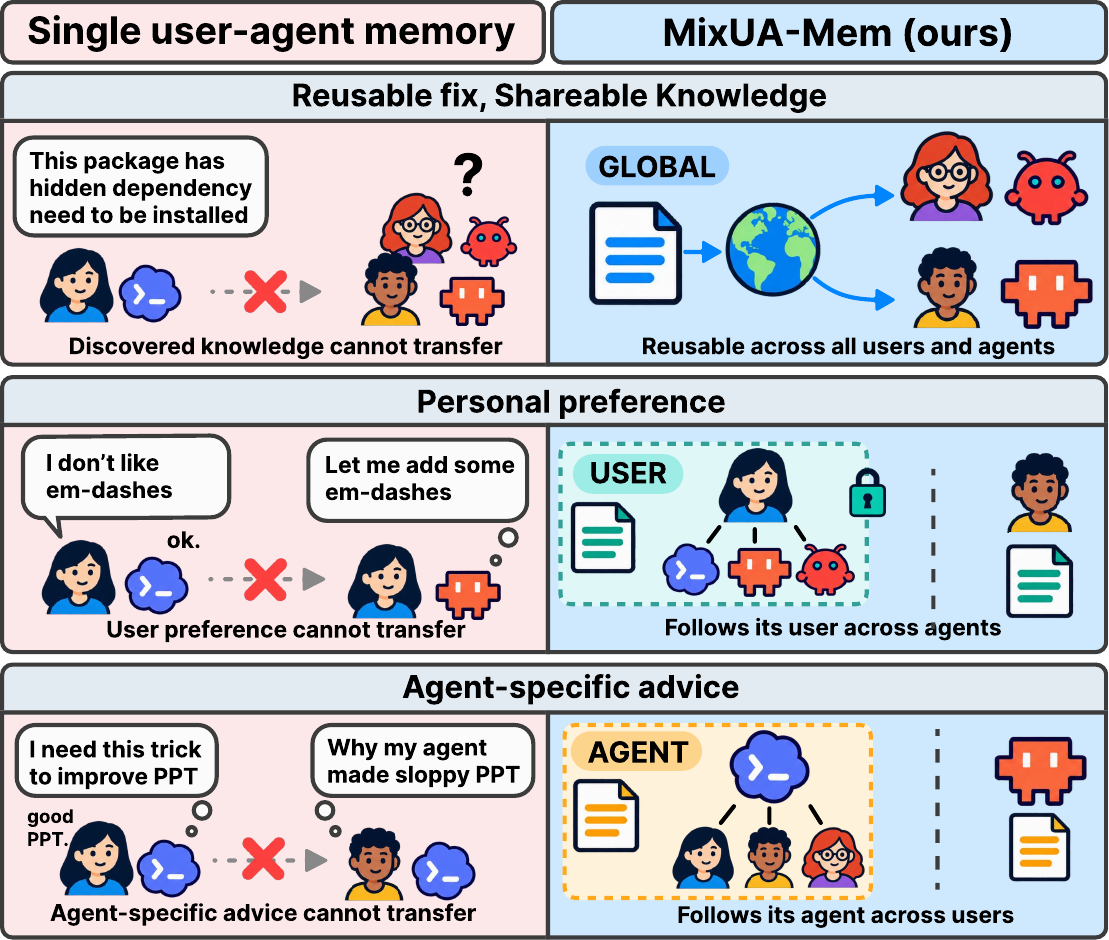}
  \caption{Illustration of what \method{} (right) adds to single user-agent memory (left).}
  \label{fig:overview}
\end{figure}

Human intelligence is distinctively \emph{cumulative}, advancing less through individual cognition than through knowledge externalized into shared media such as writing and tools, where a discovery made once accrues to the whole population \citep{tomasello2009cultural,henrich2015secret}. Recent work on self-evolving agents has begun to externalize their experience in a similar way, into skill libraries \citep{wang2023voyager,yang2026skillopt}, persistent memory \citep{packer2023memgpt}, and harness code \citep{yin2025godel,lee2026meta,ning2026code}, and group-learning systems pool these lessons across users and agents \citep{zhang2026coevoskills,ma2026skillclaw,yang2026federatedskill}. In parallel, work on mixed multi-principal agents has emerged \citep{yang2026multi,ren2026gatemem}, in which a single agent serves many users and inter-user privacy becomes a first-order concern. Yet sharing knowledge across a large population of heterogeneous users and agents is harder than simply pooling every lesson into one shared store. A lesson that helps one user or agent may not help another, and some of what an agent learns about a user should never reach others. Sharing well therefore requires deciding who can benefit from each lesson and what information must remain private.

\begin{figure*}[t]
  \centering
  \includegraphics[width=0.98\linewidth]{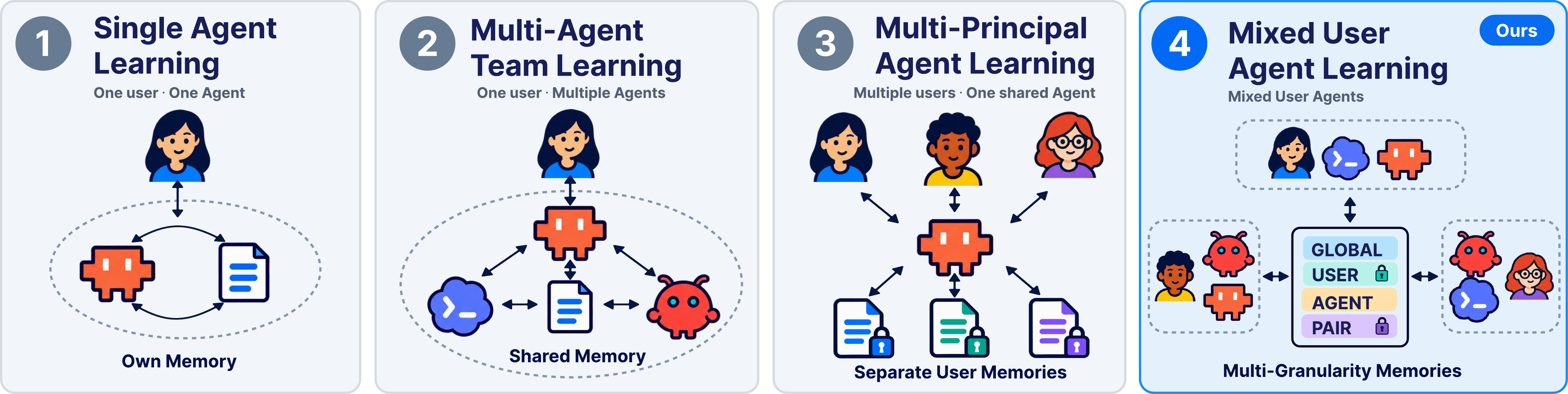}
  \caption{Four settings of agent evolution.
  \textbf{(1)}  One agent serving a single user improves from its own trajectories.
  \textbf{(2)} A fixed agent team under a single user co-evolves.
  \textbf{(3)} One agent serves many users, where inter-user privacy becomes a first-order concern.
  \textbf{(4)} A mixed population of independent users and heterogeneous agents co-evolves.}
  \label{fig:evolution-settings}
\end{figure*}

We call this setting \emph{cross-learning}, where independent users and agents learn from one another's experience.
To design such a system, we identify three core properties that learned knowledge must satisfy.
\emph{(i) Transferability.} A new insight, once discovered by any user or agent, should have the opportunity to benefit the whole population.
\emph{(ii) Audience selectivity.} Each learned item possesses an intrinsic \emph{generalization scope}, the maximal set of user-agent pairs for which it is valid (Fig.~\ref{fig:overview}). This scope may be narrow or broad, for instance, specific to a single user as with personal preferences, tied to a particular agent family as with tool-calling guides, or shared globally across the entire population as with newly discovered wiki entries.
\emph{(iii) Confidentiality.} Each learned item also carries an intrinsic \emph{disclosure boundary}, the widest scope it may occupy before it exposes information private to its origin user. A private item, such as a user's identity, should stay with its origin user, whereas a non-private item, such as a newly discovered bug in an open-source project, is safe to share across the population.

The step from self-evolving agents to self-evolving populations remains open (Fig.~\ref{fig:evolution-settings}).
Multi-agent teams improve collectively but serve one principal within one trust boundary \citep{zhang2026coevoskills,bhattacharyya2026social}.
Systems that pool experience across users fix who receives it in advance. SkillClaw \citep{ma2026skillclaw} broadcasts validated skills to everyone, FederatedSkill \citep{yang2026federatedskill} exchanges distilled skill patches between clients that each bundle one user with one agent, and Collaborative Memory \citep{rezazadeh2025collaborative} separates private from shared memory under access policies.
A lesson can generalize along two independent axes, across users and across agents. Treating each axis as either specific to the lesson's origin or open to the whole population yields four audience scopes, namely one user-agent pair, one user across agents, one agent across users, and the whole population. Existing memory systems cover only subsets of these scopes (Table~\ref{tab:related-memory-scopes}).

In this work, we introduce \method{} (Sec.~\ref{sec:method}), a cross-learning memory for mixed populations, such as a team sharing coding assistants or a platform serving many customers with heterogeneous agentic products, where each user's privacy must be protected.
We cast cross-learning as deciding who reads each lesson (Sec.~\ref{sec:Preliminary}) and give memory four scopes, by whether a lesson is user-specific, agent-specific, both, or neither (Sec.~\ref{sec:multi-granularity-memory}).
\method{} writes and reviews scoped entries, keeping private evidence with its owner (Sec.~\ref{sec:writing}), and retrieves the most relevant ones each pair may read (Sec.~\ref{sec:retrieval}).
To evaluate, we introduce \bench{} (Sec.~\ref{sec:benchmark}), modifying established agent tasks with planted user preferences and private facts in a mixed-user-agent co-learning setup, evaluating utility, personalization, and privacy together. To the best of our knowledge, \bench{} is the first benchmark to measure cross-user and cross-agent learning under privacy and preference constraints.
Our experiments (Sec.~\ref{sec:experiments}) show that (1)~\textbf{memory sharing helps beyond private memory}: four-scope memory gains up to 16 points of pass rate against 4 for memory private to each pair, and up to 14 even on tasks previously attempted only by different users with different agents; (2)~\textbf{the gain extends to new tasks}: on tasks no user-agent pair had attempted before, \method{} gains 10.1 points over no memory, nearly twice the 5.5 of a single global memory; (3)~\textbf{utility gains coexist with personalization and privacy}: \method{} retains up to 90\% of previously stated preferences, and no more than 0.02\% of probes find a planted private fact in a forbidden scope; and (4)~\textbf{store review sharpens placement}, mostly improving utility, preference retention, and privacy at once.
Our contributions are (1)~\method{}, a four-scope memory that learns where each lesson belongs and holds task knowledge, environment facts, and preferences in one store; (2)~\bench{}, a benchmark measuring utility, transfer, preference retention, privacy, and cost across heterogeneous LLM-harness combinations; and (3)~findings that the gain is experience crossing users and agents, that the four scopes divide the work, and that periodic store review sharpens this division (Sec.~\ref{sec:results}).

\section{Related work}
\label{sec:related}

\begin{table}[t]
  \centering
  \footnotesize
  \setlength{\tabcolsep}{1.6pt}
  \renewcommand{\arraystretch}{1.12}
  \begin{tabular}{@{}l@{\hspace{2pt}}cccc@{\hspace{6pt}}c@{}}
    \toprule
    System & \scriptsize P & \scriptsize U & \scriptsize A & \scriptsize G & Sharing rule \\
    \midrule
    MemGPT {\scriptsize\citep{packer2023memgpt}} & \schalf & \scnone & \scnone & \scnone & fixed \\
    A-Mem {\scriptsize\citep{NEURIPS2025_19909c36}} & \schalf & \scnone & \scnone & \scnone & fixed \\
    MIRIX {\scriptsize\citep{wang2025mirixmultiagentmemoryllmbased}} & \scnone & \schalf & \scnone & \scnone & fixed \\
    G-Memory {\scriptsize\citep{NEURIPS2025_136a45cd}} & \scnone & \scnone & \schalf & \schalf & fixed \\
    SkillClaw {\scriptsize\citep{ma2026skillclaw}} & \scnone & \scnone & \scnone & \scfull & fixed \\
    FederatedSkill {\scriptsize\citep{yang2026federatedskill}} & \schalf & \scnone & \schalf & \schalf & fixed \\
    Collab.\ Mem.\ {\scriptsize\citep{rezazadeh2025collaborative}} & \schalf & \scfull & \schalf & \schalf & policy \\
    AIM {\scriptsize\citep{johnson2026aimprivacyawareinteroperablememory}} & \scnone & \scfull & \scnone & \scfull & policy \\
    ShareMem {\scriptsize\citep{hu2026learningothersactingyou}} & \schalf & \schalf & \scnone & \schalf & fixed \\
    \midrule
    \textbf{\method{} (ours)} & \scfull & \scfull & \scfull & \scfull & \textbf{learned} \\
    \bottomrule
  \end{tabular}
  \caption{Memory systems mapped onto our four scopes from their published designs (\protect\scfull{}~explicit, \protect\schalf{}~approximate, \protect\scnone{}~none). P, U, A, and G are the \PAIR{}, \USER{}, \AGENT{}, and \GLOBAL{} (Sec.~\ref{sec:multi-granularity-memory}). \emph{Sharing rule} says whether recipients are fixed, set by policy, or learned per entry.}
  \label{tab:related-memory-scopes}
\end{table}

\paragraph{Self-evolving agents.}
Self-evolving agents turn their own trajectories into lasting improvements without retraining the base model, through test-time adaptation, reusable skills and memory, or changes to their own harness \citep{fang2025comprehensive,acikgoz2025self,wang2023voyager,lin2026muse,yin2025godel,packer2023memgpt}.
Multi-agent teams extend this loop to a fixed group under one principal \citep{zhang2026coevoskills,bhattacharyya2026social}, and recent systems pool experience across independent users \citep{ma2026skillclaw,yang2026federatedskill}.
Table~\ref{tab:related-memory-scopes} maps representative memory systems onto our four scopes.
In most work, who reads a memory is fixed by design, from memories serving one pair or user (MemGPT, A-Mem, MIRIX) to shared or federated stores (G-Memory, SkillClaw, FederatedSkill, ShareMem), while Collaborative Memory and AIM set it by access policy.

\paragraph{Agent memory and multi-user agents.}
A-Mem \citep{NEURIPS2025_19909c36} adaptively builds, links, and evolves memory notes, and G-Memory \citep{NEURIPS2025_136a45cd} retrieves a cooperating team's hierarchical experience per agent.
\method{} adds explicit ownership and audience routing, and the disclosure boundaries that independent users, unlike one team, require.
Multi-user agents, in which one agent serves several users, make access control and memory governance central \citep{yang2026multi,ren2026gatemem}, and contextual-privacy benchmarks ask whether information reaches an inappropriate recipient \citep{ruzzetti2026muppet}.
We extend these concerns to a population with many users and heterogeneous agents, and measure leakage in released evidence and stored memory.
In addition, LongMemEval \citep{wu2025longmemeval} tests extraction, multi-session and temporal reasoning, updates, and abstention in assistant memory.

\section{Method}
\label{sec:method}
\label{sec:setting}

\begin{figure*}[t]
  \centering
  \includegraphics[width=\textwidth]{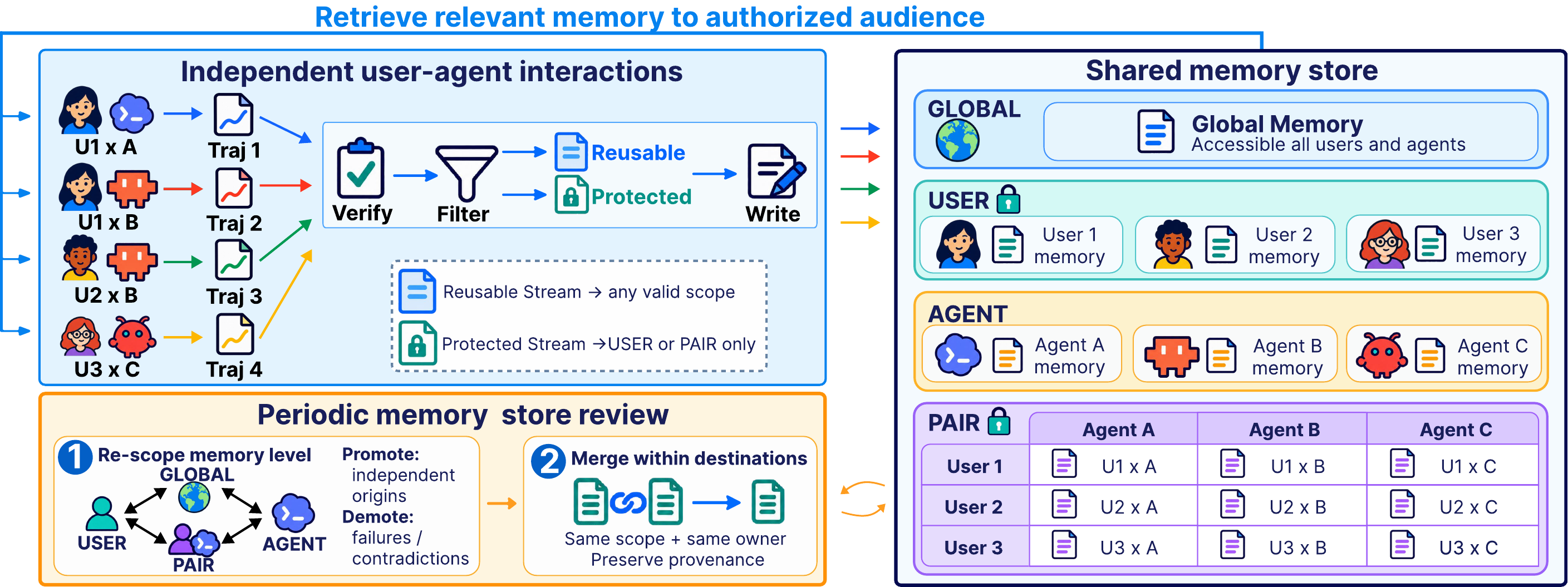}
  \caption{The cross-learning loop. Independent user-agent pairs produce trajectories, each split into a reusable and a protected stream. The memory writer places knowledge into four scopes, confining protected entries to the owner's \USER{} and \PAIR{} memory, and a periodic store review promotes, demotes, and merges entries.}
  \label{fig:framework}
\end{figure*}

\method{} turns independent trajectories into one shared memory (Fig.~\ref{fig:framework}).
We cast cross-learning as deciding who reads each entry, then describe how we write, review, and retrieve entries in four scopes (Appendix~\ref{app:situated}, Appendix~\ref{app:prompts}).

\subsection{Problem formulation}
\label{sec:Preliminary}
A population has users $\users$ and agents $\agents$, where each agent is a fixed LLM-harness configuration and a pair $(u,a)\in\users\times\agents$ is one user working with one agent.
Each attempt of a task by a pair yields a trajectory $\tau$ and a verifier outcome $y\in\{0,1\}$, and a memory writer distills task trajectories into structured memory entries $e$. Each entry has an authorized \emph{audience} $\mathrm{Aud}(e)\subseteq\users\times\agents$, the set of user-agent pairs permitted to access it.
Following Sec.~\ref{sec:intro}, an entry has a \emph{generalization scope} $\mathcal{G}(e)$, the pairs for which it is valid, and a \emph{disclosure boundary} $\mathcal{D}(e)$, the pairs it may reach without exposing its origin user.
We require
\begin{equation}
\label{eq:placement}
 \mathrm{Aud}(e)\subseteq\mathcal{G}(e)\cap\mathcal{D}(e),
\end{equation}
so that useful knowledge reaches every pair it can while private knowledge stays with its owner.

\subsection{Multi-granularity scopes}
\label{sec:multi-granularity-memory}
An entry written by pair $(u,a)$ can be kept to its user or opened to all users, and kept to its agent or opened to all agents.
The four combinations give four scopes, ordered by inclusion (Fig.~\ref{fig:scope-matrix} in Appendix~\ref{app:situated}).
\PAIR{} ($\{u\}\times\{a\}$) is read only by the user-agent pair itself, \USER{} ($\{u\}\times\agents$) by the user with any agent, \AGENT{} ($\users\times\{a\}$) by the agent with any user, and \GLOBAL{} ($\users\times\agents$) by everyone.
A pair's view $V(u,a)$ is thus its own \PAIR{} memory, its user's \USER{} memory, its agent's \AGENT{} memory, and \GLOBAL{}, never another user's \USER{} memory or another pair's \PAIR{} memory.
Only opening the user axis exposes an entry to other users, so only that step is gated on privacy (Fig.~\ref{fig:scope-matrix}).

Intuitively, each scope suits a different kind of knowledge.
\GLOBAL{} holds what helps everyone, such as a library defect or even the fix for one task, and \USER{} holds a user's preferences, such as a document format.
\AGENT{} holds advice for one agent, such as a tool-calling quirk or a mistake it tends to make, and \PAIR{} holds a user's preferences that conflict across agents, such as asking a verbose agent for terse replies but a terse one for detail.

\subsection{Memory writing}
\label{sec:writing}
In a deployed system, user-agent interactions arrive at any time and each may hold a valuable lesson, so memory should learn from every interaction and periodically review the whole store.
We simulate this in rounds (Fig.~\ref{fig:framework}). In each round, every pair attempts its tasks with the memory it may read, and the round's trajectories are then written into the store one at a time before the next round begins.

\paragraph{Dual-stream writing.}
\label{sec:situated}
After each round, we split each trajectory into a \emph{reusable} and a \emph{protected} stream and send them to two isolated writers, so that a writer that sees protected content never writes a shared entry.
We do not rely on the writer to label its own entries. Instead, we label each entry by the stream it was written from, so a protected entry can land only in its owner's \USER{} or \PAIR{} memory.

Each writer then writes memory entries based on one stream.
An entry consists of a \emph{situation} (when the lesson applies), the lesson, a scope, and provenance (the originating tasks, pairs, rounds, and outcomes).
We require every lesson to be conditioned on a situation, so that lessons stay organized and retrieval can match them to the tasks they apply to.
To test whether the design depends on how entries are produced, we implement two writers, an \emph{LLM writer} that returns all operations in one response and an \emph{agentic writer} that can inspect the sample and the store and make multi-turn edits.

\paragraph{Periodic store review.}
\label{sec:review}
Since a writer sees one trajectory at a time, it can scope a lesson too narrowly or too widely, and different trajectories can yield duplicate lessons.
We therefore add an optional periodic store review, which after every round re-scopes each entry with the whole store in view and then merges entries within each destination.

\subsection{Memory retrieval}
\label{sec:retrieval}
Before each attempt, we retrieve from the user-agent pair's view $V(u,a)$.
By default, we rank the entries of each scope by the cosine similarity between the embeddings of their situation and the task prompt, and add them to the prompt up to a per-scope character budget.
In the file-based variant, we instead place the pair's view in the container as one file per scope and instruct the agent to read or \texttt{grep} it for relevant entries.

\section{\bench{}}
\label{sec:benchmark}

\bench{} has two layers.
(1) Tasks from established benchmarks test agents' task success, and (2) a user layer tests what memory retains and leaks, which the verifiers do not measure.

\subsection{Dataset}
\label{sec:dataset}

We source tasks from seven established datasets (Appendix~\ref{app:curation}, Fig.~\ref{fig:task-topics}): AutoCodeBench \citep{chou2025autocodebench}, FeatureBench \citep{zhou2026featurebench}, SpreadsheetBench (verified) \citep{ma2024spreadsheetbench}, SWT-Bench (verified) \citep{mundler2024swtbench}, USACO \citep{shi2024usaco}, Terminal-Bench~2.0 \citep{merrill2026terminal}, and VMax Tasks \citep{vmax2026tasks}.
We modify the Harbor framework \citep{Harbor_Framework} and run tasks in Docker sandboxes, keeping each environment, instruction, and verifier unchanged and standardized.

\subsection{Privacy injection}
\label{sec:privacy-injection}

Cross-user sharing is safe only if private content never reaches a scope that other users can read. Since the source benchmarks contain no private information and do not evaluate privacy, we test privacy leakage by planting private facts into the task execution environment of existing tasks, personalized for each user based on a real-world survey.
We give each user a synthetic persona with nine always-private items of personally identifiable information (PII) and one fact per category of a privacy survey \citep{pew2014privacypanel}.
We take each fact's privacy label from per-user survey responses, so the same category can be private for one user and not for another, which forces a learning system to judge privacy on a per-user basis.
We plant these facts \emph{ambiently}, in files with neutral names at the project root of the task environment, to simulate a messy workspace in which private information is more likely to leak unexpectedly as the agent naturally explores the environment.
For example, a half-filled loan application is planted as \texttt{form.txt} (Algorithm~\ref{alg:injection}).
We then measure \emph{exposure}, how often the agent opens a planted file, and \emph{leakage}, whether a planted fact appears in the trajectory passed to the memory writer, in the user's own memory store, or in another user's memory store (Appendix~\ref{app:boundary}).

\subsection{Preference injection}
\label{sec:preference-injection}

Beyond privacy, sharing must also preserve personalization.
A memory that pools lessons across users can raise task success while forgetting how each user expects work to be done.
We similarly modify benchmark tasks by asking the model to write a summary document upon completion of each task, and give each user four document preferences: a filename, a sign-off date format, an ordered set of sections, and one banned markup, such as em dashes. A user has consistent preferences across rounds but can have conflicting preferences when using different agents.
We state each user's preferences only in round~1, and later rounds rely on memory to carry them.
For scoring, we use an automatic parser to check the written document and its adherence to each user's preferences.
Appendix~\ref{app:construction} lists the exact tasks, survey data, users, and both injection procedures, and Appendix~\ref{app:metrics} defines every metric.

\section{Experiments}
\label{sec:experiments}
\label{sec:results}

\begin{table*}[t]
\centering
\footnotesize
\setlength{\tabcolsep}{2.4pt}
\begin{tabular*}{\textwidth}{@{\extracolsep{\fill}}lrrrrrrrrrr@{}}
\toprule
 & \multicolumn{3}{c}{Utility} & \multicolumn{2}{c}{Preference} & \multicolumn{2}{c}{Privacy} & \multicolumn{3}{c}{Resource} \\
\cmidrule(lr){2-4}\cmidrule(lr){5-6}\cmidrule(lr){7-8}\cmidrule(l){9-11}
Condition & Pass R3$\uparrow$ & $\Delta{\uparrow}$ & $\Delta_{\text{cross}}{\uparrow}$ & Evol.$\uparrow$ & Held-out$\uparrow$ & Pref. leak$\downarrow$ & Exposed$\downarrow$ & Turns & Memory & Cost \\
\midrule
No memory & 72.0{\scriptsize$\pm$1.2} & --\sig{} & --\sig{} & 11 & -- & \na{} & 8 & 14.8 & 0 & 0.30 \\
\PAIR{} only & 77.6{\scriptsize$\pm$1.5} & +4.3{\scriptsize$\pm$0.8}\sig{**} & +1.6\sig{} & 80 & 79 & \na{} & 20 & 14.8 & 23.0 & 0.60 \\
\USER{} only & 80.4{\scriptsize$\pm$1.4} & +8.8{\scriptsize$\pm$0.8}\sig{**} & +3.3\sig{} & 37 & 20 & \na{} & 12 & 13.8 & 39.6 & 1.24 \\
\AGENT{} only & 78.6{\scriptsize$\pm$0.4} & +5.1{\scriptsize$\pm$1.5}\sig{**} & +3.3\sig{} & 12 & 10 & 1.2$^\ddagger$ & 9 & 13.8 & 39.7 & 0.66 \\
\quad + protected stream & 80.7{\scriptsize$\pm$0.1} & +7.3{\scriptsize$\pm$0.3}\sig{**} & +5.4\sig{*} & 33 & 18 & 26.3 & 10 & 13.4 & 39.0 & 1.05 \\
\GLOBAL{} only & 85.5{\scriptsize$\pm$1.6} & +11.5{\scriptsize$\pm$0.6}\sig{**} & +8.9\sig{**} & 11 & 10 & 0.4$^\ddagger$ & 6 & 11.4 & 39.7 & 1.77 \\
\quad + protected stream & 86.9{\scriptsize$\pm$0.1} & +13.2{\scriptsize$\pm$0.9}\sig{**} & +12.6\sig{**} & 15 & 12 & 5.5 & 8 & 11.5 & 39.6 & 3.70 \\
\midrule
\method{} (LLM) & 88.1{\scriptsize$\pm$1.7} & +15.3{\scriptsize$\pm$1.8}\sig{**} & +13.0\sig{**} & \textbf{90} & \textbf{90} & 4.8 & 26 & 13.2 & 60.2 & 3.35 \\
\method{} (LLM) + review & \textbf{89.8{\scriptsize$\pm$1.3}} & \textbf{+16.3{\scriptsize$\pm$1.1}\sig{**}} & \textbf{+13.9\sig{**}} & 89 & \textbf{90} & 3.6 & 24 & 13.9 & 61.4 & 3.43 \\
\quad 10k char retrieval limit & 85.3{\scriptsize$\pm$1.4} & +13.1{\scriptsize$\pm$1.0}\sig{**} & +10.9\sig{**} & 89 & 78 & 3.0 & 17 & 13.0 & 22.8 & 3.31 \\
\quad file-based memory & 87.2{\scriptsize$\pm$2.3} & +14.5{\scriptsize$\pm$1.8}\sig{**} & +12.9\sig{**} & 82 & 75 & --$^\dagger$ & 18 & 17.0 & 608$^\dagger$ & 3.58 \\
\method{} (agentic) & 86.3{\scriptsize$\pm$2.0} & +12.8{\scriptsize$\pm$2.7}\sig{**} & +12.4\sig{**} & 80 & 78 & 1.3 & 11 & 12.9 & 47.7 & 1.24 \\
\method{} (agentic) + review & 87.1{\scriptsize$\pm$0.4} & +14.0{\scriptsize$\pm$1.3}\sig{**} & +10.7\sig{**} & 86 & 84 & \textbf{0.9} & 12 & 12.6 & 48.0 & 1.22 \\
\bottomrule
\end{tabular*}
\caption{Main results, mean{\scriptsize$\pm$}s.d.\ over three runs. $\Delta$ is the gain over no memory in rounds~2-3 (pp), and $\Delta_{\text{cross}}$ the same on tasks tried before only by pairs sharing neither user nor agent. Preference is retention (\%). Privacy is the \% of probes finding a user's preference in another user's memory and of attempts opening a planted file. Resource is per-task memory in k characters and cost including memory writing. $^\dagger$files searched on demand, whose leak rate is not comparable; $^\ddagger$no protected stream, so the preference leak is not comparable.}
\label{tab:overall-results}
\end{table*}

\begin{figure*}[t]
\centering
\includegraphics[width=\textwidth]{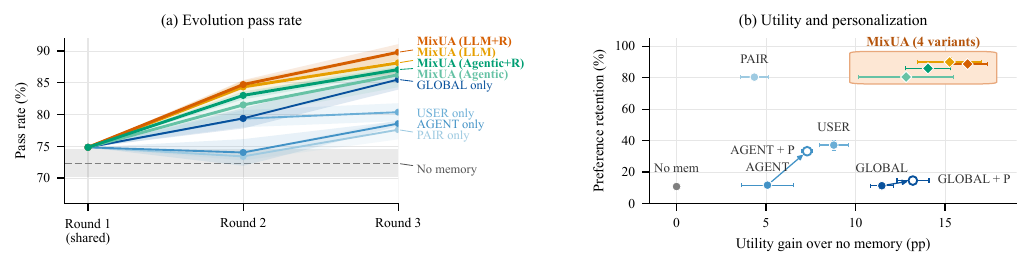}
\caption{(a) Pass rate by evolution round, mean $\pm$ s.d.\ over runs (+R denotes store review); no memory cannot learn, so it is one band. (b) Utility gain against preference retention; only the four \method{} variants attain both. Hollow markers (+P) are the \emph{+ protected stream} rows of Table~\ref{tab:overall-results}, with arrows from the rows they modify.}
\label{fig:main}
\end{figure*}

\subsection{Experimental setup}
\label{sec:experiment-setup}
\label{sec:evaluation-protocol}

\begin{figure*}[t]
\centering
\includegraphics[width=\textwidth]{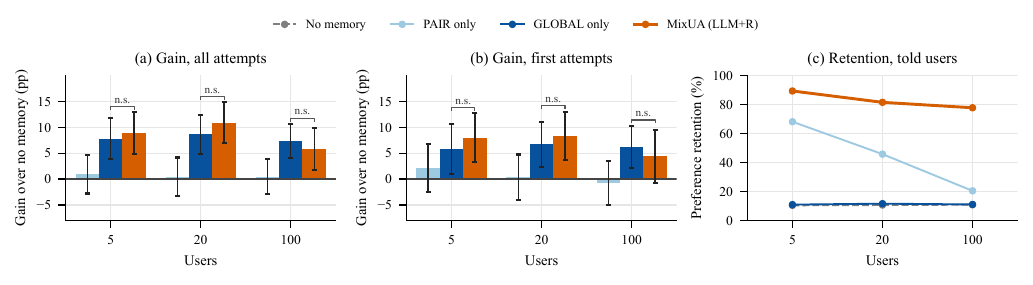}
\caption{Number of users under random task assignment, one seed each. (a, b) Gain over no memory in rounds~2-8, 95\% intervals; n.s.: \method{} vs.\ \GLOBAL{}-only. (c) Preference retention of users told in round~1.}
\label{fig:scale}
\end{figure*}

We pair synthetic users with seven LLM-harness combinations (Claude Code with Opus~5 or Sonnet~5, Hermes and OpenClaw with Sonnet~5, and Codex with GPT-5.6 Sol, Luna, or Terra).
We compare no memory, four single-scope stores (\GLOBAL{}-only, \USER{}-only, \AGENT{}-only, and \PAIR{}-only), and \method{} with an LLM or a Codex agent as memory writer, each with and without store review (Table~\ref{tab:condition-settings}).
Importantly, a single-scope store writes only its one scope but is not limited to that scope's usual kind of knowledge, and it can record every lesson its audience may read. \GLOBAL{}-only, for example, can keep a lesson specific to one user or agent by naming them in the entry (\emph{when user~A is working on [situation]}). Similarly, the \USER{}-only store can write pair-style entries by mentioning the specific agent.
Since no existing system evaluates a mixed user-agent population on general-purpose agent tasks, these stores also stand in for memory design in prior work: \PAIR{}-only for an assistant memory such as MemGPT, \USER{}-only for a personal memory such as MIRIX, \AGENT{}-only for a team memory such as G-Memory, and \GLOBAL{}-only for a broadcast skill library such as SkillClaw (Table~\ref{tab:related-memory-scopes}).
We further ablate what each scope adds by removing it from \method{}, which leaves combinations of the remaining scopes (Sec.~\ref{sec:ablations}).
The memory writer uses Claude Opus~5, and retrieval uses Titan Text Embeddings~v2 (Appendices~\ref{app:situated} and~\ref{app:repro}).
We test scheduled task assignment (Table~\ref{tab:overall-results}), random task assignment with five users (Fig.~\ref{fig:stream}), and a user-scaling experiment with 20 and 100 users (Fig.~\ref{fig:scale}).

\paragraph{Scheduled tasks.}
An \emph{attempt} is one user-agent pair running one task, and several pairs may attempt the same task during evolution. To control who can learn from whom, we fix a schedule of 171 attempts per round that rotates tasks across pairs for a total of three rounds, so that a task is met again by the same pair, by a pair sharing its user or agent, or by a pair sharing neither, and we can tell whose experience produced each gain (Appendix~\ref{app:curation}).
We then freeze each store and run 20 held-out tasks (140 attempts) with learning off, masking which memory is visible (none, each scope alone, all scopes, a shuffled user's scopes, or the preferences restated as a ceiling).
We repeat each condition three times at a 40k-character retrieval budget per scope, and rerun \method{} (LLM) + review at 1k-20k characters per scope to test whether the gain is merely more context.

\paragraph{Random tasks.}
The scheduled assignment above relies on repeating earlier tasks. To get closer to a real deployment, we also evaluate under \emph{random task assignment} for eight rounds, where each round draws 60 tasks from a 546-task pool and gives each to a random pair. It tests whether memory also helps on a task that no pair has attempted, where only lessons learned on other tasks can help. We run it for no memory, \PAIR{}-only, \GLOBAL{}-only, and \method{} (LLM) + review at 20k characters per scope (Appendix~\ref{app:stream}).

\paragraph{Metrics.}
We report three main metrics: (1) \emph{utility}, the pass rate under each source benchmark's verifier; (2) \emph{preference retention}, the rate of satisfied document checks after the round-1 statement is withdrawn; and (3) \emph{privacy}, the share of memory probes finding a planted private fact in a forbidden scope or a user's preference in another user's memory (Appendix~\ref{app:boundary}).
Gains are in percentage points (pp), with exact McNemar significance marked $^{*}$\,$p<0.05$ and $^{**}$\,$p<0.01$.
Appendix~\ref{app:metrics} defines each metric, and Table~\ref{tab:costs} adds context size, cost, and efficiency.

\subsection{Main results}
\label{sec:main-results}

\begin{figure*}[t]
\centering
\includegraphics[width=\textwidth]{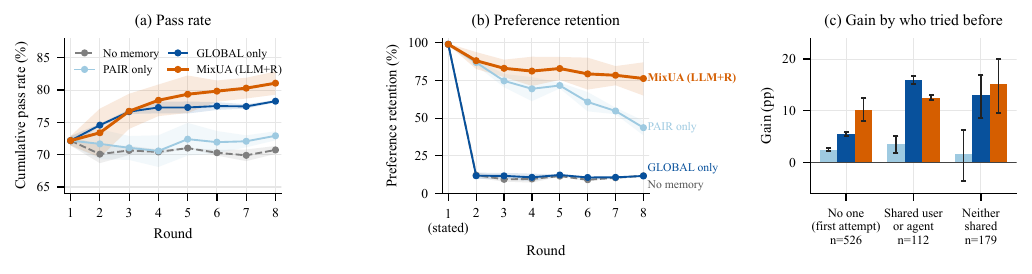}
\caption{Random task assignment, two seeds pooled; shading and whiskers span the two seeds. (a) Cumulative pass rate. (b) Retention of the preferences stated in round~1. (c) Gain over no memory by who tried the task before.}
\label{fig:stream}
\end{figure*}

\begin{figure}[t]
\centering
\includegraphics[width=\columnwidth]{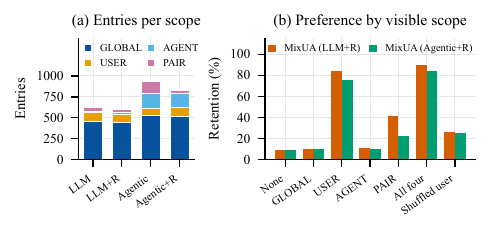}
\caption{Each scope holds different knowledge. (a) \method{} store composition after round~3. (b) Held-out preference retention by visible scopes.}
\label{fig:placement}
\end{figure}

\begin{table}[t]
\centering
\footnotesize
\setlength{\tabcolsep}{3.2pt}
\begin{tabular}{@{}lrrrrr@{}}
\toprule
 & \multicolumn{2}{c}{Evolution} & \multicolumn{3}{c}{Held-out} \\
\cmidrule(lr){2-3}\cmidrule(l){4-6}
Budget & Context & $\Delta$ & Pass & Pref. & Cost \\
\midrule
1k & 5.7 & +8.4 & 71.2 & 60.3 & 54 \\
5k & 13.2 & +10.4 & 70.3 & 66.1 & 54 \\
10k & 22.8 & +13.1 & 71.6 & 78.0 & 60 \\
20k & 37.8 & +14.0 & 75.4 & 87.3 & 65 \\
40k & 61.4 & +16.3 & 75.4 & 90.9 & 81 \\
\midrule
\GLOBAL{} only (40k) & 39.7 & +11.5 & -- & -- & -- \\
Empty store & -- & -- & 63.6 & 8.2 & 45 \\
\bottomrule
\end{tabular}
\caption{Retrieval budget per memory scope for \method{} (LLM) + review. Evolution columns give the context per task (k chars) and the gain (pp). Cost in USD. More details in Fig.~\ref{fig:budget} and Table~\ref{tab:budget}.}
\label{tab:budget-main}
\end{table}

\paragraph{Scheduled tasks.}
\label{sec:r-main}
\label{sec:r-transfer}
\label{sec:r-privacy}
Each single-scope store trades utility for preferences (Fig.~\ref{fig:main}b, Table~\ref{tab:overall-results}): \GLOBAL{}-only gains the most (11.5 points) but retains no more preferences than no memory (11\%), and \PAIR{}-only the reverse (4.3 points, 80\%).
\method{} attains both, gaining 13-16 points while retaining 80-90\%.
This gain is not merely due to task repetition (Fig.~\ref{fig:transfer}a).
On tasks tried before by a different user and agent, where only the \GLOBAL{} scope can carry the lesson, \method{} gains \mbox{11-14} points against 8.9 for \GLOBAL{}-only and 1.6 for \PAIR{}-only, and when an earlier pair has the same user or agent, which the \USER{} and \AGENT{} scopes cover, \method{} gains up to 20 against 14.7.
Every agent but Claude Code-Opus, which is already near ceiling, gains 11-23 points (Fig.~\ref{fig:transfer}b).
Sharing also keeps planted facts private. Although agents with \method{} open a planted file in 11-26\% of attempts, three of its four variants place none in a forbidden scope in about 18k probes each, and the memory store review removes the agentic writer's 4 violations.
Preferences leak less than in \GLOBAL{} and \AGENT{}-only with the protected stream (\mbox{0.9-4.8\%} vs.\ \mbox{5.5-26.3\%}).
Store review improves both writers, raising $\Delta$ from 15.3 to 16.3 points with the LLM writer and from 12.8 to 14.0 with the agentic writer, lifting the agentic writer's preference retention from 80\% to 86\%, and lowering preference leaks from 4.8\% to 3.6\% and from 1.3\% to 0.9\%.

\paragraph{Random tasks.}
\label{sec:r-stream}
In rounds~2-8, 526 of 821 attempts are on a task that no user-agent pair had attempted before. On these, \method{} gains 10.1 points ($p<10^{-7}$), nearly twice the 5.5 of \GLOBAL{}-only, with the gap widening over rounds (Fig.~\ref{fig:stream}a,c).
On previously attempted tasks the two gain about equally, so the added scopes help most where no task-specific lesson exists.
Preferences stated in round~1 persist (Fig.~\ref{fig:stream}b). \method{} retains 76\% in round~8, whereas \PAIR{}-only declines from 87\% to 44\%, plausibly because its preferences compete with 1.7k task entries for one budget.

\subsection{Ablations}
\label{sec:ablations}

\begin{table}[t]
\centering
\footnotesize
\setlength{\tabcolsep}{2.6pt}
\begin{tabular}{@{}lrrrlrr@{}}
\toprule
 & \multicolumn{3}{c}{Benchmark} & \multicolumn{3}{c}{Scope probe} \\
\cmidrule(lr){2-4}\cmidrule(l){5-7}
Removed & $\Delta{\uparrow}$ & $\Delta_{\text{cross}}{\uparrow}$ & Pref.$\uparrow$ & Metric & With & W/o \\
\midrule
None & 15.3 & 13.0 & 90 & & & \\
\GLOBAL{} & \textbf{9.0} & \textbf{4.6} & 90 & & & \\
\USER{} & 10.5 & 8.8 & \textbf{81} & & & \\
\AGENT{} & 14.8 & 12.5 & 91 & Crowd pass$\uparrow$ & 89 & \textbf{40} \\
 & & & & Spillover$\downarrow$ & 44 & \textbf{67} \\
\PAIR{} & 15.2 & 12.3 & 93 & Leak$\downarrow$ & 22 & \textbf{95} \\
\bottomrule
\end{tabular}
\caption{What each scope adds: \method{} (LLM) without one scope. Benchmark columns as in Table~\ref{tab:overall-results}; scope-probe columns in \% (Appendix~\ref{app:extra-ablations}). \textbf{Bold} means worst.}
\label{tab:ablation-snapshot}
\end{table}

\paragraph{What each scope adds.}
To evaluate what each memory scope contributes, we test scope combinations by removing one scope at a time from \method{} (LLM), so that removing \GLOBAL{}, for example, leaves \USER{}, \AGENT{}, and \PAIR{} (Table~\ref{tab:ablation-snapshot}).
Importantly, as in the single-scope baselines, the remaining scopes can still record what the missing scope would have held. For example, without \AGENT{}, a \GLOBAL{} entry could read \emph{when agent~X builds \dots}.
On utility, removing \GLOBAL{} causes the largest drop, cutting $\Delta$ from 15.3 to 9.0 points and $\Delta_{\text{cross}}$ from 13.0 to 4.6.
On preferences, removing \USER{} causes the largest drop (90\% to 81\%).
Although all four scopes together are best or within run-to-run noise of the best on every measure, we find that removing \AGENT{} or \PAIR{} changes relatively little.
We therefore build these situations as scope probes (Appendix~\ref{app:extra-ablations}).
\emph{Crowd pass} is the share of build probes passed when each agent needs its own build mode and every scope's retrieval budget fits about one entry. \emph{Spillover} is, at the normal budget, the share of probes whose memory shows another agent's mode; and \emph{leak} is the share of a user's tasks on their other agents whose memory shows a work context the user asked to keep with one agent.
Without \AGENT{}, crowd pass falls from 89\% to 40\% and spillover rises from 44\% to 67\%; without \PAIR{}, leak rises from 22\% to 95\%.
Both scopes are therefore needed, where \AGENT{} keeps an agent's own lessons retrievable and away from other agents, and \PAIR{} keeps a user's context with the agent it was meant for.

\paragraph{Held-out scope masking.}
\label{sec:r-placement}
\label{sec:r-review}
To check that each scope holds the knowledge it is meant to, we freeze each evolved store and, on the 20 held-out tasks, show the agent one scope, all scopes, or a shuffled user's memory (Fig.~\ref{fig:placement}).
Although most entries are \GLOBAL{} task lessons, \GLOBAL{} or \AGENT{} alone retains preferences no better than an empty store (8-12\%), so a user's conventions live in \USER{} and \PAIR{}.
All scopes together retain 84-90\% after review, near the 99\% of restating the preferences, whereas a shuffled user's memory retains about a quarter at the same pass rate.

\paragraph{Robustness to memory reading and writing.}
\label{sec:budget}
\label{sec:r-files}
\label{sec:r-split}
We next ask whether the performance gain comes from the four level scopes or merely because the four-scope store writes more text.
Varying the retrieval budget rules this out (Table~\ref{tab:budget-main}). We find that at 10k characters per scope, \method{} delivers 23k per task, under 60\% of \GLOBAL{}-only, yet still gains more (13.1 versus 11.5 points), and on held-out tasks 20k per scope matches the 40k pass rate at about 20\% lower cost, though preference retention keeps rising with the budget.
Replacing retrieval with searchable files, as coding harnesses commonly do, gains nearly as much (14.5 points), but agents take more turns (17.0 versus 13.9) and retain fewer preferences (75\% versus 90\%).

\section{Conclusion}
We introduced \bench{}, a benchmark for cross-learning among heterogeneous users and agents, and \method{}, a memory system that assigns memory entry to the scope whose pairs it benefits while protecting privacy.
Our results show that knowledge shared across users and agents improves performance beyond what any single user-agent pair gets in isolation, and this improvement relies on the complementary roles of the four scopes. As humans and agents increasingly coexist at scale, intelligence becomes collective, and knowing where a lesson belongs may matter as much as learning it.

\section*{Limitations}
\label{sec:limitations}

\paragraph{Independent users and agents.}
We model users as independent principals and agents as independent configurations, so every private fact belongs to exactly one user and every disclosure boundary stops at that user.
In reality, users are related and organized in hierarchies, from teams to departments to whole organizations, each sharing its own projects, conventions, and secrets, as with a project secret shared within one team. Agents are similar: sharing could happen among base models, harnesses, versions, and even role-playing personas.
Knowledge owned by a team is neither pair-level private nor global, which our privacy boundary rule cannot express.
Modeling relations and hierarchies among users and agents, with sharing and disclosure boundaries at each level, is future work.

\paragraph{Audiences between one and all.}
Even within our model, an audience is one user or all users, and one agent or all agents, so a lesson valid for a subset, such as one team or the agents built on one model, must sit either at \USER{} or \AGENT{}, where it reaches too few, or at \GLOBAL{}, where it reaches too many.
Group scopes need a way to discover or declare groups, and every audience between one and all adds a disclosure boundary that the privacy gate must judge.
We leave group scopes, and a richer lattice of audiences that keeps every readership auditable, to future work.

\paragraph{Scale of the population.}
We envision an open co-learning system that anyone can join, but our study has 5, 20, or 100 synthetic users, seven agents, and at most 546 tasks over eight rounds.
At global scale the store grows without bound, many lessons compete for one budgeted context, and a review that reads the whole store no longer fits in one pass.
At a larger scale, memory writing could incur a large cost.
In addition, real users bring richer needs and preference requirements.
Scaling the store, retrieval, and review of cross-learning to an open population with real users, and bounding the writing cost as the store grows, are future work.

\section*{Ethical Considerations}
All users and personal facts in this study are synthetic, which gives us known probe targets without collecting real personal records.
A deployed shared memory would also need consent before a user's experience is shared, a way for users to inspect and delete what was learned from them, and defenses against contributors who write misleading lessons on purpose; our design records provenance, which makes inspection and deletion possible, but we do not evaluate these mechanisms.
Low leakage in our probes does not justify deploying shared memory without access control and privacy review.

\bibliography{ref}
\clearpage
\appendix
\raggedbottom
\section{Additional Results and Findings}
\label{app:extra}
Beyond Sec.~\ref{sec:experiments}, this appendix reports the full results of scheduled task assignment (Sec.~\ref{app:extra-scheduled}), the design and results of random task assignment (Sec.~\ref{app:stream}), additional ablations of the four scopes (Sec.~\ref{app:extra-ablations}), and three case studies (Sec.~\ref{sec:cases}).

\subsection{Scheduled task assignment}
\label{app:extra-scheduled}
\begin{figure*}[tp]
\centering
\includegraphics[width=\textwidth]{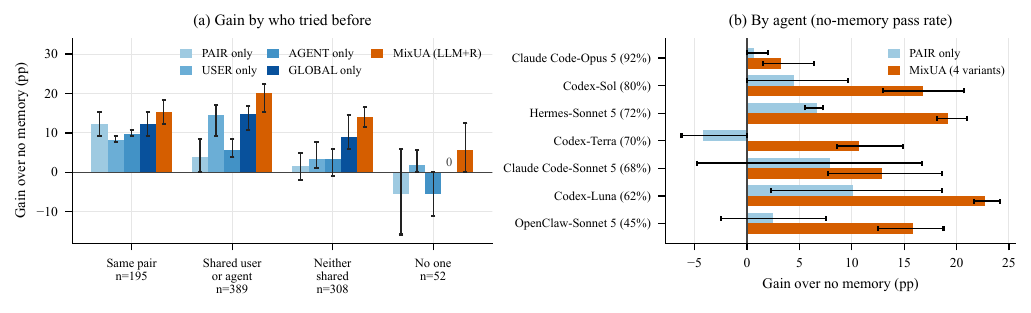}
\caption{Where the gain comes from, as gain over no memory in rounds~2-3. (a) By who tried the task before; bars pool three runs, whiskers span runs, and $n$ counts attempts. (b) By agent, ordered by no-memory pass rate; whiskers span runs.}
\label{fig:transfer}
\end{figure*}

\begin{figure}[tp]
\centering
\includegraphics[width=\columnwidth]{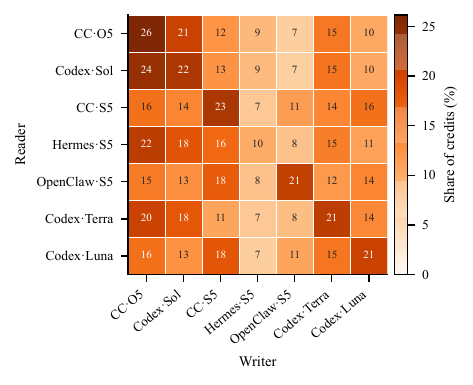}
\caption{Origin of the memory each agent reads, as the share of origin credits on entries delivered to each reader by the \method{} LLM stores, rounds~2-3. Entries are matched to the store snapshot that served them; a merged entry credits every origin.}
\label{fig:provenance}
\end{figure}

\begin{figure*}[tp]
\centering
\includegraphics[width=\textwidth]{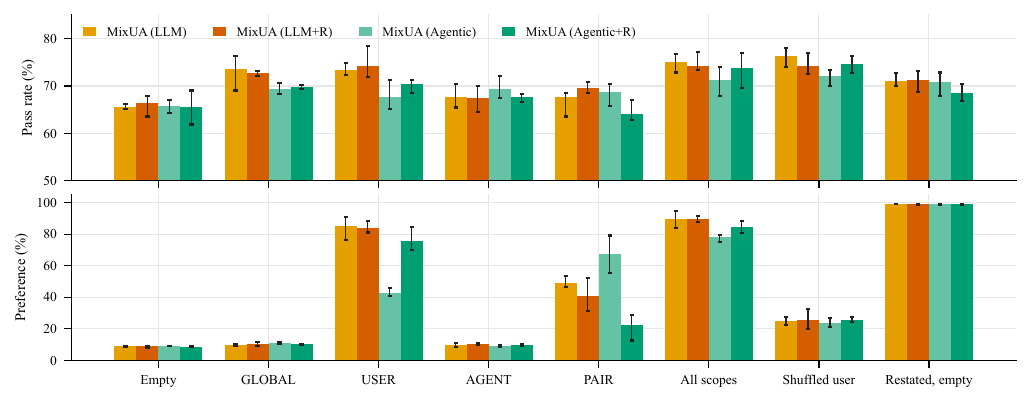}
\caption{Held-out pass rate (top) and preference retention (bottom) of the \method{} stores under each mask; bars are means over the three runs, whiskers span single passes.}
\label{fig:heldout}
\end{figure*}

\begin{figure*}[tp]
\centering
\includegraphics[width=\textwidth]{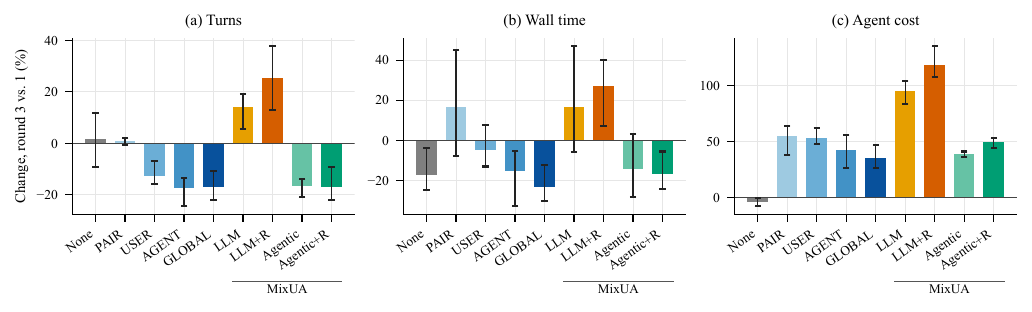}
\caption{Round-3 vs.\ round-1 efficiency on repeated tasks passed every time; negative values mean that round~3 needed fewer turns, less time, or lower agent cost. Bars are means over runs, whiskers span runs.}
\label{fig:efficiency}
\end{figure*}

\begin{figure*}[tp]
\centering
\includegraphics[width=\textwidth]{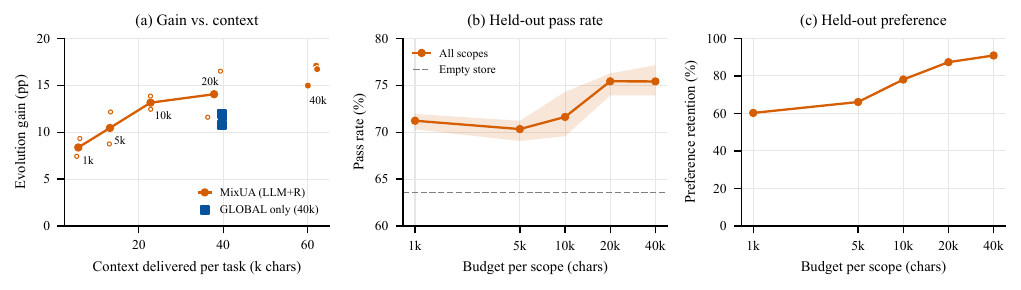}
\caption{Retrieval budget for \method{} (LLM) + review. (a) Evolution gain against context delivered per task. (b, c) Held-out pass rate and preference retention when only the retrieval budget changes; shading spans three passes.}
\label{fig:budget}
\end{figure*}

Fig.~\ref{fig:transfer} splits the gain by who tried the task before and by agent, Tables~\ref{tab:exposure} and~\ref{tab:agent-gains} give these gains in full, and Fig.~\ref{fig:provenance} shows whose memory each agent reads.
Tables~\ref{tab:heldout} and~\ref{tab:ood} and Fig.~\ref{fig:heldout} report every held-out arm, Table~\ref{tab:costs} and Fig.~\ref{fig:efficiency} the resources, and Fig.~\ref{fig:budget} and Table~\ref{tab:budget} the retrieval budget.

\paragraph{Reading the held-out arms.}
The held-out stage complements the evolution stage rather than replacing it.
Evolution is itself an online test, since every attempt is graded before the store learns from it, and there \method{} (LLM) + review gains the most of any condition, both overall (16.3 points) and on tasks tried before only by pairs sharing neither user nor agent (13.9; Tables~\ref{tab:overall-results} and~\ref{tab:exposure}).
The held-out stage then freezes each store after round~3 and, on 20 tasks that no pair met during evolution, asks two further questions: whether the memory still helps with learning off, and, by masking scopes, where each kind of knowledge lives.

On the first, \method{} keeps the utility of a shared store and adds the user's preferences to it (Table~\ref{tab:heldout}).
With all scopes visible, its LLM store gains the most of any store (9.4 points, against 8.0 for \GLOBAL{}-only and 6.2 for \PAIR{}-only), and every \method{} store retains 78-90\% of the preferences, against 10\% for \GLOBAL{}-only (Table~\ref{tab:overall-results}).
Each single-scope store is again strong on at most one side: \GLOBAL{}-only comes closest on pass rate but retains no more than an empty store, and \PAIR{}-only comes closest on retention (79\%) but gains less.
Restating the preferences to an empty store raises pass rate by 3.1-5.0 points, and three of the four \method{} stores gain 2.5-4.5 points more than that, so the learned memory adds to what the stated preferences alone give.
The gain also reaches datasets the receiving agent never met during evolution, where only other agents' experience can help (Table~\ref{tab:ood}): three of the four \method{} stores gain significantly there (6.9-8.6 points), and no single-scope store does (\GLOBAL{}-only 6.6, \PAIR{}-only 3.0).

On the second, the masks recover the division of labor that the four scopes are designed for (Fig.~\ref{fig:heldout}).
\GLOBAL{} or \AGENT{} alone retains preferences no better than an empty store (8-12\%), while \USER{} alone retains up to 91\% and all scopes together 76-95\%.
Handing every scoped entry to the wrong owner (\emph{shuffled user}) keeps the pass rate but cuts retention to 21-33\%, so the task knowledge helps whichever user receives it, while each user's conventions help only their owner and stay in that user's own \USER{} and \PAIR{} memory.
Together, the two stages show a gain that is earned online, persists with learning off, and reaches datasets the receiving agent never met, with each user's preferences kept with that user.

\begin{table*}[tp]
\centering
\small
\begin{tabular}{@{}lrrrr@{}}
\toprule
Condition & Same pair & Shared user/agent & Neither shared & Unseen \\
 & $n$=195 & $n$=389 & $n$=308 & $n$=52 \\
\midrule
\PAIR{} only & +12.3\sig{**} & +3.9\sig{*} & +1.6\sig{} & -5.6\sig{} \\
\USER{} only & +8.2\sig{**} & +14.4\sig{**} & +3.3\sig{} & +1.8\sig{} \\
\AGENT{} only & +9.7\sig{**} & +5.7\sig{**} & +3.3\sig{} & -5.5\sig{} \\
\GLOBAL{} only & +12.3\sig{**} & +14.7\sig{**} & +8.9\sig{**} & +0.0\sig{} \\
\midrule
\method{} (LLM) & +13.3\sig{**} & +19.3\sig{**} & +13.0\sig{**} & \textbf{+5.8\sig{}} \\
\method{} (LLM) + review & \textbf{+15.4\sig{**}} & \textbf{+20.0\sig{**}} & \textbf{+13.9\sig{**}} & +5.7\sig{} \\
\method{} (agentic) & +14.9\sig{**} & +14.1\sig{**} & +12.4\sig{**} & -1.9\sig{} \\
\method{} (agentic) + review & +14.4\sig{**} & +18.2\sig{**} & +10.7\sig{**} & +1.9\sig{} \\
\bottomrule
\end{tabular}
\caption{Gain over no memory (pp) on attempts in rounds~2-3, by who attempted the task in an earlier round; three runs pooled. Bold marks the best per column.}
\label{tab:exposure}
\end{table*}

\begin{table*}[tp]
\centering
\small
\begin{tabular}{@{}lrrrrrr@{}}
\toprule
Agent & None (\%) & \PAIR{} & \USER{} & \AGENT{} & \GLOBAL{} & \method{} \\
\midrule
OpenClaw-Sonnet-5 & 45.0 & +2.5\sig{} & +13.3\sig{**} & +5.0\sig{} & +15.4\sig{**} & \textbf{+15.8}\sig{**} \\
Codex-Luna & 62.0 & +10.1\sig{*} & +16.4\sig{**} & +10.9\sig{**} & +21.3\sig{**} & \textbf{+22.7}\sig{**} \\
Claude Code-Sonnet-5 & 68.2 & +7.9\sig{*} & +6.3\sig{*} & +8.5\sig{*} & +9.3\sig{**} & \textbf{+12.9}\sig{**} \\
Codex-Terra & 69.8 & -4.2\sig{} & +1.0\sig{} & +4.2\sig{} & +10.4\sig{**} & \textbf{+10.7}\sig{**} \\
Hermes-Sonnet-5 & 72.0 & +6.7\sig{*} & +9.6\sig{**} & +4.3\sig{} & +6.0\sig{*} & \textbf{+19.2}\sig{**} \\
Codex-Sol & 80.1 & +4.5\sig{} & +10.3\sig{**} & +5.1\sig{} & +16.0\sig{**} & \textbf{+16.8}\sig{**} \\
Claude Code-Opus-5 & 92.1 & +0.7\sig{} & +3.3\sig{} & -1.3\sig{} & \textbf{+4.0}\sig{} & +3.3\sig{**} \\
\bottomrule
\end{tabular}
\caption{Per-agent no-memory pass rate and gain (pp) on rounds~2-3, pooled over runs; \method{} pools its four variants. Bold marks the largest gain per agent.}
\label{tab:agent-gains}
\end{table*}

\begin{table*}[tp]
\centering
\footnotesize
\setlength{\tabcolsep}{3pt}
\begin{tabular*}{\textwidth}{@{\extracolsep{\fill}}llrrrrrrr@{}}
\toprule
 & & \multicolumn{2}{c}{Run 1} & \multicolumn{2}{c}{Run 2} & \multicolumn{2}{c}{Run 3} & Pooled \\
\cmidrule(lr){3-4}\cmidrule(lr){5-6}\cmidrule(lr){7-8}\cmidrule(l){9-9}
Store & Visible & Pass & Pref. & Pass & Pref. & Pass & Pref. & $\Delta$ (gain/loss) \\
\midrule
\method{} (LLM) & Empty & 66.2 & 9 & 65.5 & 9 & 65.2 & 9 & -- \\
 & GLOBAL & 69.1 & 10 & 75.4 & 9 & 76.3 & 10 & +7.8\sig{**} (42/10) \\
 & USER & 73.4 & 76 & 74.8 & 89 & 72.3 & 91 & +8.2\sig{**} (53/19) \\
 & AGENT & 65.5 & 8 & 67.1 & 11 & 70.5 & 9 & +2.2\sig{} (21/12) \\
 & PAIR & 66.1$^{\times2}$ & 46 & 68.3 & 48 & 68.6 & 54 & +1.2\sig{} (29/24) \\
 & All scopes & 75.4$^{\times3}$ & 84 & 72.9 & 90 & 76.8 & 95 & +9.4\sig{**} (57/18) \\
 & Shuffled user & 74.5$^{\times2}$ & 27 & 78.1 & 25 & 76.3 & 23 & \textbf{+10.8}\sig{**} (57/13) \\
 & Restated, empty & 72.8 & 99 & 70.0 & 99 & 70.6 & 99 & +4.9\sig{*} (40/20) \\
\midrule
\method{} (LLM) + review & Empty & 63.6 & 8 & 67.6 & 9 & 67.9 & 9 & -- \\
 & GLOBAL & 72.1 & 12 & 73.2 & 9 & 73.2 & 10 & +6.5\sig{**} (41/14) \\
 & USER & 78.4 & 83 & 72.5 & 88 & 72.0 & 81 & \textbf{+8.3}\sig{**} (52/18) \\
 & AGENT & 64.5 & 11 & 70.1 & 10 & 67.9 & 11 & +0.7\sig{} (21/18) \\
 & PAIR & 69.4$^{\times2}$ & 52 & 68.6 & 31 & 70.8 & 40 & +2.9\sig{} (28/16) \\
 & All scopes & 75.4$^{\times3}$ & 91 & 73.9 & 91 & 73.4 & 88 & +7.5\sig{**} (49/18) \\
 & Shuffled user & 75.2$^{\times2}$ & 21 & 72.5 & 24 & 75.0 & 33 & +7.3\sig{**} (49/19) \\
 & Restated, empty & 73.2 & 99 & 68.7 & 99 & 71.9 & 99 & +5.0\sig{**} (36/16) \\
\midrule
\method{} (agentic) & Empty & 67.1 & 9 & 64.3 & 9 & 66.2 & 9 & -- \\
 & GLOBAL & 69.3 & 10 & 68.3 & 11 & 70.7 & 12 & +3.6\sig{*} (31/16) \\
 & USER & 66.7 & 41 & 65.2 & 41 & 71.2 & 46 & +2.2\sig{} (31/22) \\
 & AGENT & 67.4 & 8 & 68.3 & 9 & 72.1 & 10 & +3.4\sig{} (31/17) \\
 & PAIR & 66.9$^{\times2}$ & 67 & 70.4 & 79 & 68.6 & 55 & +1.9\sig{} (24/16) \\
 & All scopes & 73.0$^{\times2}$ & 76 & 67.9 & 79 & 72.7 & 79 & +5.0\sig{**} (35/14) \\
 & Shuffled user & 73.3$^{\times2}$ & 21 & 70.0 & 27 & 72.9 & 22 & \textbf{+6.2}\sig{**} (36/10) \\
 & Restated, empty & 72.9 & 99 & 72.1 & 98 & 67.9 & 99 & +4.9\sig{**} (35/15) \\
\midrule
\method{} (agentic) + review & Empty & 61.9 & 8 & 69.1 & 8 & 65.5 & 9 & -- \\
 & GLOBAL & 70.0 & 11 & 70.3 & 10 & 69.4 & 10 & +4.6\sig{**} (31/12) \\
 & USER & 71.2 & 70 & 71.2 & 72 & 68.6 & 84 & +4.8\sig{*} (40/20) \\
 & AGENT & 68.3 & 10 & 67.9 & 11 & 66.7 & 9 & +1.9\sig{} (26/18) \\
 & PAIR & 66.3$^{\times2}$ & 28 & 62.9 & 27 & 63.0 & 13 & -1.2\sig{} (19/24) \\
 & All scopes & 74.6$^{\times2}$ & 88 & 69.6 & 80 & 77.0 & 85 & +7.5\sig{**} (46/15) \\
 & Shuffled user & 74.8$^{\times2}$ & 27 & 72.7 & 24 & 76.3 & 26 & \textbf{+9.2}\sig{**} (48/10) \\
 & Restated, empty & 70.5 & 99 & 66.9 & 99 & 68.3 & 99 & +3.1\sig{} (35/22) \\
\midrule
\PAIR{} only & Empty & 66.4 & 8 & 65.5 & 9 & 66.2 & 8 & -- \\
 & All scopes & 76.3$^{\times3}$ & 82 & 67.6 & 76 & 70.0 & 78 & \textbf{+6.2}\sig{**} (39/13) \\
\midrule
\USER{} only & Empty & 66.9 & 9 & 69.8 & 8 & 67.7 & 10 & -- \\
 & All scopes & 70.5$^{\times3}$ & 22 & 73.0 & 17 & 71.5 & 20 & \textbf{+2.5}\sig{} (22/12) \\
\midrule
\AGENT{} only & Empty & 66.9 & 9 & 66.2 & 9 & 68.9 & 9 & -- \\
 & All scopes & 69.9$^{\times3}$ & 10 & 71.2 & 10 & 73.5 & 11 & \textbf{+4.4}\sig{**} (29/11) \\
\midrule
\GLOBAL{} only & Empty & 67.2 & 9 & 65.0 & 8 & 66.9 & 9 & -- \\
 & All scopes & 74.6$^{\times3}$ & 11 & 74.5 & 10 & 74.8 & 10 & \textbf{+8.0}\sig{**} (47/14) \\
\bottomrule
\end{tabular*}
\caption{Held-out arms on the 20 held-out tasks (140 attempts per arm), with learning off. Each block is one store frozen after round~3, and its rows mask what the agent sees: nothing (\emph{Empty}), one scope, all scopes, all scopes with every \USER{}, \PAIR{}, and \AGENT{} entry handed to another owner (\emph{Shuffled user}; \GLOBAL{} entries have no owner), or only the restated preferences (\emph{Restated, empty}). A single-scope store has only an empty and an all-scope arm, and the all-scope rows compare stores. Pass rate and preference retention in \%; $\times k$ marks the mean of $k$ repeated passes, and the restated arm reports stated compliance. $\Delta$ is against the store's empty arm on matched attempts, pooled over the three runs, with the attempts gained and lost in parentheses; bold marks each store's largest gain.}
\label{tab:heldout}
\end{table*}

\begin{table*}[tp]
\centering
\small
\begin{tabular*}{\textwidth}{@{\extracolsep{\fill}}lrrrrrrrr@{}}
\toprule
 & \multicolumn{4}{c}{In-domain ($n\approx250$)} & \multicolumn{4}{c}{Out-of-domain ($n\approx168$)} \\
\cmidrule(lr){2-5}\cmidrule(l){6-9}
Store & Empty & All & \GLOBAL{} & Restated & Empty & All & \GLOBAL{} & Restated \\
\midrule
\method{} (LLM) & 62.9 & \textbf{+9.3\sig{**}} & +8.9\sig{**} & +7.2\sig{**} & 69.6 & \textbf{+8.6\sig{*}} & +6.2\sig{} & +2.5\sig{} \\
\method{} (LLM) + review & 62.9 & +6.6\sig{*} & +7.9\sig{**} & +6.1\sig{**} & 71.4 & +6.9\sig{*} & +6.9\sig{} & +3.1\sig{} \\
\method{} (agentic) & 64.9 & +3.0\sig{} & +0.8\sig{} & +4.7\sig{} & 67.3 & +8.5\sig{*} & +7.9\sig{*} & +4.9\sig{} \\
\method{} (agentic) + review & 61.8 & +9.2\sig{**} & +5.9\sig{**} & +6.3\sig{*} & 70.8 & +5.4\sig{} & +3.0\sig{} & -1.2\sig{} \\
\midrule
\PAIR{} only & 62.8 & +8.5\sig{**} & --\sig{} & --\sig{} & 70.8 & +3.0\sig{} & --\sig{} & --\sig{} \\
\USER{} only & 64.0 & +3.3\sig{} & --\sig{} & --\sig{} & 74.3 & +1.2\sig{} & --\sig{} & --\sig{} \\
\AGENT{} only & 65.4 & +5.0\sig{*} & --\sig{} & --\sig{} & 70.1 & +3.6\sig{} & --\sig{} & --\sig{} \\
\GLOBAL{} only & 63.4 & +9.1\sig{**} & --\sig{} & --\sig{} & 70.7 & +6.6\sig{} & --\sig{} & --\sig{} \\
\bottomrule
\end{tabular*}
\caption{Held-out arms split by domain, pooled over the three runs. Out-of-domain attempts are held-out tasks from a dataset the receiving agent never met during evolution, so only other agents' experience with that dataset can reach it. \emph{Empty} is the pass rate (\%), and the other columns give the gain over the empty arm (pp) with all scopes, \GLOBAL{}-only, or an empty store with restated preferences. Bold marks the largest all-scope gain.}
\label{tab:ood}
\end{table*}

\begin{table*}[tp]
\centering
\small
\begin{tabular*}{\textwidth}{@{\extracolsep{\fill}}lrrrrrrr@{}}
\toprule
 & Context & Ungraded & \multicolumn{2}{c}{USD per run} & \multicolumn{3}{c}{R3 vs.\ R1, always-pass tasks (\%)} \\
\cmidrule(lr){4-5}\cmidrule(l){6-8}
Condition & (k chars) & (\%) & Agent & Memory & Turns & Time & Cost \\
\midrule
No memory & 0 & 1.7 & 154 & 0 & +2 & -18 & -4 \\
\PAIR{} only & 23.0 & 1.4 & 175 & 101 & +1 & +16 & +55 \\
\USER{} only & 39.6 & 1.8 & 180 & 346 & -13 & -5 & +53 \\
\AGENT{} only & 39.7 & 1.6 & 180 & 118 & \textbf{-18} & -15 & +43 \\
\GLOBAL{} only & 39.7 & 1.4 & 156 & 580 & -17 & \textbf{-23} & \textbf{+36} \\
\midrule
\method{} (LLM) & 60.2 & 1.2 & 190 & 1,168 & +14 & +17 & +96 \\
\method{} (LLM) + review & 61.4 & 0.7 & 193 & 1,191 & +25 & +27 & +118 \\
\method{} (agentic) & 47.7 & 1.4 & 181 & 418 & -17 & -14 & +39 \\
\method{} (agentic) + review & 48.0 & 0.9 & 174 & 418 & -17 & -17 & +50 \\
\bottomrule
\end{tabular*}
\caption{Resources. \emph{Context} is characters delivered per task in rounds~2-3, \emph{Ungraded} the errors and timeouts in rounds~2-3, and \emph{USD} the evolution-stage agent and memory-writing spend per run. Efficiency columns give the change from round~1 to round~3 on (pair, task) combinations passed in all three rounds (20-25 per condition and run); bold marks the largest reduction.}
\label{tab:costs}
\label{tab:efficiency}
\end{table*}

\begin{table*}[tp]
\centering
\small
\begin{tabular*}{\textwidth}{@{\extracolsep{\fill}}lrrrrrrrr@{}}
\toprule
 & \multicolumn{5}{c}{Re-evolved store} & \multicolumn{3}{c}{Fixed store, held-out} \\
\cmidrule(lr){2-6}\cmidrule(l){7-9}
Budget & Runs & Context & $\Delta$ & Pass R3 & Pref. R3 & Pass & Pref. & Cost \\
\midrule
1k & 2 & 5.7 & +8.4 & 83.3 & 59.5 & 71.2{\scriptsize$\pm$0.8} & 60.3{\scriptsize$\pm$0.4} & \textbf{54{\scriptsize$\pm$2}} \\
5k & 2 & 13.2 & +10.4 & 87.2 & 73.7 & 70.3{\scriptsize$\pm$1.1} & 66.1{\scriptsize$\pm$0.1} & 54{\scriptsize$\pm$1} \\
10k & 2 & 22.8 & +13.1 & 85.3 & 87.5 & 71.6{\scriptsize$\pm$2.4} & 78.0{\scriptsize$\pm$0.5} & 60{\scriptsize$\pm$3} \\
20k & 2 & 37.8 & +14.0 & 85.1 & 86.9 & \textbf{75.4{\scriptsize$\pm$1.3}} & 87.3{\scriptsize$\pm$0.3} & 65{\scriptsize$\pm$2} \\
40k & 3 & 61.4 & \textbf{+16.3} & \textbf{89.8} & \textbf{89.9} & 75.4{\scriptsize$\pm$1.6} & \textbf{90.9{\scriptsize$\pm$0.6}} & 81{\scriptsize$\pm$2} \\
\bottomrule
\end{tabular*}
\caption{Retrieval budget per scope for \method{} (LLM) + review. \emph{Runs} counts evolution runs, whose means the next four columns give, \emph{Context} is k characters delivered per task, and $\Delta$ is the evolution gain over no memory (pp). Held-out columns report the all-scope arm of an evolved store with only the retrieval budget changed, mean{\scriptsize$\pm$}s.d.\ over three passes, with cost in USD per 140-attempt arm. Bold marks the best per column.}
\label{tab:budget}
\end{table*}

\subsection{Random task assignment}
\label{app:stream}

\paragraph{Design.}
The pinned schedule repeats a round-1 task in 149 of 171 round-2 attempts and 154 of 171 round-3 attempts, so it measures transfer mostly on tasks that some pair has met before.
Random task assignment draws every round afresh, sampling 60 distinct tasks from a 546-task pool (the training split of the seven datasets, minus ungradeable tasks) and giving each to one of the 35 pairs at random, so a task returns in a later round only by chance and almost never to the same pair.
All four conditions face the same task sequence, with round~1 executed once and replayed to the others, so task difficulty cancels in every paired comparison.
We run two seeds, each fixing every draw and assignment, for eight rounds (480 attempts per condition and seed), with the same users, agents, writer, privacy split, and evaluations as scheduled task assignment and a store review after every round, but at 20k characters per scope, the budget that reaches the 40k held-out pass rate (Sec.~\ref{sec:budget}), and without a held-out stage, since first attempts are held out along the whole curve.
Per condition, 1.4-2.0\% of the attempts in rounds~2-8 are ungraded.

\begin{figure*}[tp]
\centering
\includegraphics[width=\textwidth]{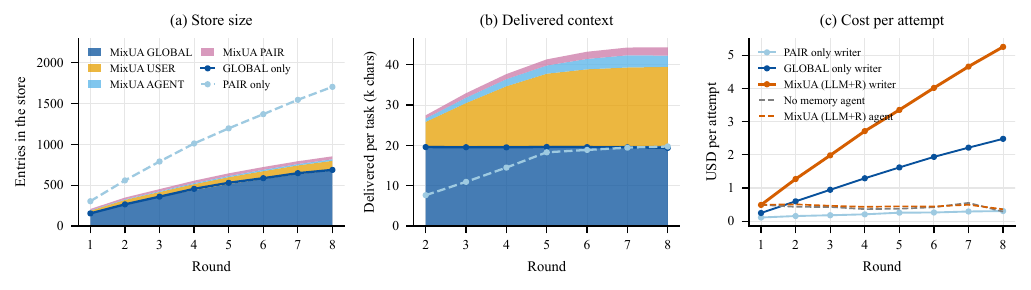}
\caption{Store dynamics under random task assignment, mean over the two seeds. (a) Entries after each round's writing and review, by scope for \method{} (LLM) + review. (b) Memory delivered per task; every scope has a 20k-character budget, which the \GLOBAL{} scope fills from round~2. (c) Writer and agent USD per attempt.}
\label{fig:stream-dynamics}
\end{figure*}

\paragraph{Results.}
Table~\ref{tab:stream} summarizes each condition, Table~\ref{tab:stream-datasets} splits the gain by dataset, Fig.~\ref{fig:stream-dynamics} shows how the stores grow, and Table~\ref{tab:scale} varies the number of users (Fig.~\ref{fig:scale}).
Over rounds~2-8, \method{} passes 82.3\% of attempts against 70.5\% without memory (11.4 points, $p<10^{-14}$), and both sharing conditions need fewer turns than no memory while \PAIR{}-only does not.
The \GLOBAL{} scope of \method{} grows like \GLOBAL{}-only (680 and 688 entries after round~8) and fills its budget from round~2, while its \USER{} memory grows to 120 entries and nearly fills its budget by round~6, so its delivered context rises from 27k to 44k characters per task.

\begin{table*}[tp]
\centering
\footnotesize
\setlength{\tabcolsep}{2.4pt}
\begin{tabular*}{\textwidth}{@{\extracolsep{\fill}}lrrrrrrrrrrrr@{}}
\toprule
 & & \multicolumn{4}{c}{Gain over no memory (pp)} & \multicolumn{2}{c}{Preference (\%)} & \multicolumn{2}{c}{Privacy (\%)} & \multicolumn{3}{c}{Resource} \\
\cmidrule(lr){3-6}\cmidrule(lr){7-8}\cmidrule(lr){9-10}\cmidrule(l){11-13}
Condition & Pass & All & First & Shared & Neither & R2-8 & R8 & Pref. leak & Opened & Turns & Memory & Cost \\
\midrule
 & & $n$=821 & $n$=526 & $n$=112 & $n$=179 & & & & & & & \\
\midrule
No memory & 70.5 & --\sig{} & --\sig{} & --\sig{} & --\sig{} & 11 & 12 & \na{} & 10 & 18.2 & 0 & 0.41 \\
\PAIR{} only & 73.0 & +2.4\sig{} & +2.5\sig{} & +3.5\sig{} & +1.7\sig{} & 66 & 44 & \na{} & 15 & 18.7 & 15.6 & 0.70 \\
\GLOBAL{} only & 79.1 & +8.5\sig{**} & +5.5\sig{**} & \textbf{+15.9\sig{**}} & +13.0\sig{**} & 11 & 12 & 0.2 & 9 & 14.8 & 19.5 & 1.94 \\
\method{} (LLM) + review & \textbf{82.3} & \textbf{+11.4\sig{**}} & \textbf{+10.1\sig{**}} & +12.5\sig{**} & \textbf{+15.1\sig{**}} & \textbf{81} & \textbf{76} & 5.8 & 27 & 16.9 & 38.7 & 3.77 \\
\bottomrule
\end{tabular*}
\caption{Random task assignment, rounds~2-8, two seeds pooled. \emph{Pass} is the pass rate (\%), and gains are over no memory on matched attempts, for all attempts and split by who attempted the task in an earlier round, either no one (\emph{First}), a pair sharing the user or the agent (\emph{Shared}), or only pairs sharing neither (\emph{Neither}); $n$ counts the attempts of \method{}. \emph{Preference} is retention over rounds~2-8 and in round~8. \emph{Privacy} gives the \% of probes finding a user's preference in another user's memory and the \% of attempts opening a planted file. \emph{Resource} gives per-attempt turns, memory as k characters delivered, and USD for the agent and the writer. Bold marks the best per column; \na{} means no cross-user scope. \GLOBAL{}-only receives no protected stream, so its preference leak is not comparable.}
\label{tab:stream}
\end{table*}

\begin{table*}[tp]
\centering
\small
\begin{tabular*}{\textwidth}{@{\extracolsep{\fill}}lrrrrrrrrrr@{}}
\toprule
 & \multicolumn{5}{c}{First attempt} & \multicolumn{5}{c}{Attempted before} \\
\cmidrule(lr){2-6}\cmidrule(l){7-11}
Dataset & $n$ & None & \PAIR{} & \GLOBAL{} & \method{} & $n$ & None & \PAIR{} & \GLOBAL{} & \method{} \\
\midrule
AutoCodeBench & 77 & 61 & +0.0\sig{} & +0.0\sig{} & +2.6\sig{} & 30 & 63 & +0.0\sig{} & +23.3\sig{*} & +20.0\sig{*} \\
FeatureBench & 85 & 61 & +10.5\sig{*} & +23.3\sig{**} & +22.4\sig{**} & 42 & 60 & -4.8\sig{} & +28.6\sig{**} & +23.8\sig{**} \\
SpreadsheetBench & 70 & 83 & -5.8\sig{} & -4.3\sig{} & +4.3\sig{} & 47 & 77 & +0.0\sig{} & +6.5\sig{} & +6.4\sig{} \\
SWT-Bench & 83 & 63 & +7.2\sig{} & +2.4\sig{} & +16.9\sig{**} & 48 & 54 & +6.4\sig{} & +19.1\sig{*} & +16.7\sig{} \\
Terminal-Bench~2.0 & 55 & 67 & +3.7\sig{} & +7.4\sig{} & +7.3\sig{} & 21 & 67 & +0.0\sig{} & +4.5\sig{} & +4.8\sig{} \\
USACO & 81 & 84 & -1.2\sig{} & -2.4\sig{} & +3.7\sig{} & 53 & 85 & +1.9\sig{} & +0.0\sig{} & +7.5\sig{} \\
VMax Tasks & 75 & 81 & +1.3\sig{} & +10.7\sig{*} & +10.7\sig{*} & 54 & 76 & +9.3\sig{} & +16.7\sig{**} & +16.7\sig{**} \\
\midrule
\textit{All} & 526 & 71 & +2.5\sig{} & +5.5\sig{**} & +10.1\sig{**} & 295 & 70 & +2.4\sig{} & +13.9\sig{**} & +13.9\sig{**} \\
\bottomrule
\end{tabular*}
\caption{Gain over no memory (pp) under random task assignment by dataset, rounds~2-8, two seeds pooled, on tasks that no pair had attempted before (\emph{First attempt}) and on tasks that some pair had (\emph{Attempted before}). \emph{None} is the no-memory pass rate (\%), and $n$ counts the matched attempts of \method{}.}
\label{tab:stream-datasets}
\end{table*}

\begin{table*}[tp]
\centering
\footnotesize
\setlength{\tabcolsep}{2.4pt}
\begin{tabular*}{\textwidth}{@{\extracolsep{\fill}}llrrrrrrrrrr@{}}
\toprule
 & & & \multicolumn{2}{c}{Gain (pp)} & \multicolumn{2}{c}{Preference (\%)} & \multicolumn{2}{c}{Privacy} & \multicolumn{3}{c}{Resource} \\
\cmidrule(lr){4-5}\cmidrule(lr){6-7}\cmidrule(lr){8-9}\cmidrule(l){10-12}
Users & Condition & Pass & All & First & Told & All & Pref. leak (\%) & Private facts & Memory & \USER{} & Cost \\
\midrule
5 & No memory & 71.3 & --\sig{} & --\sig{} & 10 & 10 & \na{} & \na{} & 0 & -- & 0.35 \\
 & \PAIR{} only & 72.3 & +1.0\sig{} & +2.2\sig{} & 68 & 68 & \na{} & \na{} & 15.5 & -- & 0.65 \\
 & \GLOBAL{} only & 79.0 & +7.8\sig{**} & +5.8\sig{*} & 11 & 11 & 0.2 & 0/5685 & 19.7 & -- & 1.89 \\
 & \method{} (LLM) + review & 80.3 & +8.9\sig{**} & +8.0\sig{**} & 89 & 89 & 3.8 & 2/9753 & 39.1 & 15.5 & 3.84 \\
\midrule
20 & No memory & 70.8 & --\sig{} & --\sig{} & 11 & 11 & \na{} & \na{} & 0 & -- & 0.38 \\
 & \PAIR{} only & 71.4 & +0.5\sig{} & +0.4\sig{} & 46 & 42 & \na{} & \na{} & 6.8 & -- & 0.52 \\
 & \GLOBAL{} only & 79.6 & +8.7\sig{**} & +6.7\sig{**} & 12 & 12 & 0.2 & 0/28097 & 19.8 & -- & 1.82 \\
 & \method{} (LLM) + review & 81.8 & +10.9\sig{**} & +8.4\sig{**} & 82 & 74 & 3.5 & 3/57913 & 30.4 & 6.3 & 3.56 \\
\midrule
100 & No memory & 72.0 & --\sig{} & --\sig{} & 11 & 11 & \na{} & \na{} & 0 & -- & 0.41 \\
 & \PAIR{} only & 72.1 & +0.5\sig{} & -0.7\sig{} & 20 & 15 & \na{} & \na{} & 3.8 & -- & 0.50 \\
 & \GLOBAL{} only & 79.4 & +7.3\sig{**} & +6.2\sig{**} & 11 & 11 & 0.1 & 0/92160 & 19.7 & -- & 1.96 \\
 & \method{} (LLM) + review & 78.1 & +5.8\sig{**} & +4.4\sig{} & 78 & 45 & 3.0 & 0/268215 & 24.8 & 2.2 & 3.39 \\
\bottomrule
\end{tabular*}
\caption{Number of users under random task assignment, rounds~2-8, one seed per size (5 users is seed~1 of Table~\ref{tab:stream}). \emph{Pass} is the pass rate (\%), and gains are over no memory on matched attempts, for all attempts and for tasks no pair had attempted before (\emph{First}). \emph{Preference} is retention over rounds~2-8, among the users whose preferences were stated in round~1 (\emph{Told}) and among all users (\emph{All}). \emph{Privacy} gives the \% of probes finding a user's preference in another user's delivered memory, and the planted private facts found over those probed. \emph{Resource} gives the k characters of memory delivered per task, in total and from the \USER{} scope, and USD per attempt for the agent and the writer. \na{} means no cross-user scope. \GLOBAL{}-only receives no protected stream, so its preference leak is not comparable.}
\label{tab:scale}
\end{table*}

\subsection{Additional ablations}
\label{app:extra-ablations}

\paragraph{Scope combinations.}
Table~\ref{tab:combo} removes scopes from \method{} (LLM) and lets the remaining scopes simulate them: the writer files a missing scope's lessons in the nearest remaining scope under the same privacy rule, with the missing condition written into the situation. Without \AGENT{}, a lesson becomes a \GLOBAL{} entry such as \emph{when agent~X builds \dots}; without \PAIR{}, a \USER{} entry such as \emph{when this user works with agent~X \dots}.
Each row thus shows what its missing scope adds beyond what the others can simulate, a stricter test than the single-scope baselines of Table~\ref{tab:overall-results}, which four scopes beat outright.
Removing \GLOBAL{} costs the most utility ($\Delta$ 9.0 instead of 15.3, $\Delta_{\text{cross}}$ 4.6 instead of 13.0), and removing \USER{} costs both utility ($\Delta$ 10.5) and preference retention (81\% instead of 90\%); every combination that keeps \USER{} retains 89-94\%, against 11\% for \GLOBAL{}-only.
\AGENT{} and \PAIR{} exist for situations in which agents need conflicting lessons or a user's context must stay with one agent, which the scope probes below build directly.

\begin{table*}[tp]
\centering
\footnotesize
\setlength{\tabcolsep}{2.4pt}
\begin{tabular*}{\textwidth}{@{\extracolsep{\fill}}lrrrrrrrrrr@{}}
\toprule
 & \multicolumn{3}{c}{Utility} & \multicolumn{2}{c}{Preference} & \multicolumn{2}{c}{Privacy} & \multicolumn{3}{c}{Resource} \\
\cmidrule(lr){2-4}\cmidrule(lr){5-6}\cmidrule(lr){7-8}\cmidrule(l){9-11}
Scopes & Pass R3$\uparrow$ & $\Delta{\uparrow}$ & $\Delta_{\text{cross}}{\uparrow}$ & Evol.$\uparrow$ & Held-out$\uparrow$ & Pref. leak$\downarrow$ & Exposed$\downarrow$ & Turns & Memory & Cost \\
\midrule
\method{} (LLM) & 88.1{\scriptsize$\pm$1.7} & +15.3{\scriptsize$\pm$1.8}\sig{**} & +13.0\sig{**} & 90 & 90 & 4.8 & 26 & 13.2 & 60.2 & 3.35 \\
\midrule
\GLOBAL{} + \PAIR{} & 87.3{\scriptsize$\pm$2.2} & +12.8{\scriptsize$\pm$1.4}\sig{**} & +11.2\sig{**} & 81 & 84 & 1.3 & 17 & 12.3 & 43.6 & 3.55 \\
\USER{} + \AGENT{} + \PAIR{} & 81.6{\scriptsize$\pm$0.4} & +9.0{\scriptsize$\pm$1.8}\sig{**} & +4.6\sig{} & 90 & 91 & 4.4 & 31 & 16.3 & 56.7 & 1.26 \\
\GLOBAL{} + \AGENT{} + \PAIR{} & 84.8{\scriptsize$\pm$2.2} & +10.5{\scriptsize$\pm$1.3}\sig{**} & +8.8\sig{**} & 81 & 81 & 0.8 & 12 & 12.4 & 45.6 & 3.37 \\
\GLOBAL{} + \USER{} + \PAIR{} & 86.8{\scriptsize$\pm$0.2} & +14.8{\scriptsize$\pm$0.6}\sig{**} & +12.5\sig{**} & 91 & 89 & 3.1 & 25 & 13.6 & 57.3 & 3.69 \\
\GLOBAL{} + \USER{} + \AGENT{} & 89.4{\scriptsize$\pm$0.0} & +15.2{\scriptsize$\pm$0.4}\sig{**} & +12.3\sig{**} & 93 & 94 & 4.4 & 18 & 12.4 & 68.6 & 3.67 \\
\bottomrule
\end{tabular*}
\caption{Scope combinations. The top row is \method{} (LLM) with all four scopes, from Table~\ref{tab:overall-results} (three runs); the others are the same system with only the scopes listed (two runs each, paired with the no-memory attempts of all three runs). Columns as in Table~\ref{tab:overall-results}.}
\label{tab:combo}
\end{table*}

\paragraph{What \AGENT{} and \PAIR{} add.}
Table~\ref{tab:scope-levels} builds the two situations and compares \method{} with the matching three-scope rows of Table~\ref{tab:combo}: without \PAIR{} is \GLOBAL{}+\USER{}+\AGENT{}, and without \AGENT{} is \GLOBAL{}+\USER{}+\PAIR{}.
In both, the conditioned entries that stand in for the missing scope fall short.

When a user asks one agent to keep their work codename, host, and ticket prefix with it, \PAIR{} is the only scope that can.
Without it, the user's one private scope is \USER{}, which all of that user's agents read, so no wording of the entry keeps it from the others: work details reach the user's personal tasks on their other agents in 95\% of cases, against 22\% with \PAIR{}, at no gain in recall (80\% against 85\%).

When one agent's build mode is wrong for the others and only an earlier verifier reveals it, \AGENT{} gives that lesson its own audience and budget.
In a crowded store, with about one entry per scope (1.6k characters) as a stand-in for the stores of a large population (under random task assignment, \GLOBAL{} already fills its budget from round~2; Fig.~\ref{fig:stream-dynamics}b), \method{} still passes 89\% of build probes and never uses another agent's mode.
Without \AGENT{}, the one \GLOBAL{} entry delivered is often a generic lesson or another agent's, so the pass rate falls to 40\% and 14\% of builds use another agent's mode.
At the normal 40k budget, \AGENT{} still raises the pass rate (93\% against 85\%, and 78\% against 56\% for users new to the lesson) and shows the other agents less of it (44\% against 67\% of prompts, with a third of the characters).
Where these situations arise, the conditioned entries that stand in for the two scopes fail: \PAIR{} keeps a user's context with one agent, and \AGENT{} keeps an agent's lesson retrievable and away from the agents it would mislead.

\begin{table}[tp]
\centering
\small
\setlength{\tabcolsep}{6pt}
\renewcommand{\arraystretch}{1.1}
\begin{tabular}{@{}llcc@{}}
\toprule
Setting & Metric & With & Without \\
\midrule
\PAIR{}-Work & Leak\,$\downarrow$ & \textbf{22} & 95 \\
 & Recall\,$\uparrow$ & \textbf{85} & 80 \\
\midrule
\AGENT{}-Crowd & Pass\,$\uparrow$ & \textbf{89} & 40 \\
 & Misapplied\,$\downarrow$ & \textbf{0} & 14 \\
\midrule
\AGENT{}-Norm & Spillover\,$\downarrow$ & \textbf{44} & 67 \\
 & Pass\,$\uparrow$ & \textbf{93} & 85 \\
\bottomrule
\end{tabular}
\caption{Scope probes: what \PAIR{} and \AGENT{} add in the situations they exist for, rounds~2-3 (\%), with three agents (Claude Code-Opus, Codex-Sol, OpenClaw) and four users. \emph{With} is \method{} (LLM); \emph{Without} drops the setting's scope, which makes it the matching three-scope row of Table~\ref{tab:combo}. \PAIR{}-Work (5~runs): a user asks one agent to keep their work codename, host, and ticket prefix with it; \emph{Leak} is the share of the user's personal tasks on their other agents whose memory shows any of them ($n$=80), and \emph{Recall} the share of work probes that recover all three ($n$=40). \AGENT{}-Crowd (2~runs) and \AGENT{}-Norm (3~runs): a build mode that only an earlier verifier reveals is right for one agent and wrong for the others, with a per-scope budget of 1.6k characters (about one entry) or the normal 40k; \emph{Pass} is the share of build probes passed, \emph{Misapplied} the share built with another agent's mode, and \emph{Spillover} the share whose memory shows another agent's mode ($n$=72 and 108). Better value in bold.}
\label{tab:scope-levels}
\end{table}

\subsection{Case studies}
\label{sec:cases}
We trace three cases through the stores of \method{} (LLM) + review (users as in Table~\ref{tab:users}).

\paragraph{A lesson crosses both boundaries.}
In round~2, Codex-Sol solved an AutoCodeBench Dart task for u4 whose specification asks a file-backed class to raise IO exceptions, and the writer stored a \GLOBAL{} lesson that file calls already throw \texttt{FileSystemException}, which implements Dart's \texttt{IOException} interface, so the exception should propagate rather than be wrapped or redeclared.
In round~3, the task went to Codex-Luna for u0, a pair sharing neither user nor agent.
Without memory, Luna failed all three runs on the same compile error, ``The class `IOException' is abstract and can't be instantiated''; with \method{} it received the lesson and passed all three, in one run choosing \texttt{FileSystemException} ``since Dart does not define a separate \texttt{IOException} class''.

\paragraph{Conventions settle where they hold.}
User u0 wants the same filename, sections, and prose-only style from every agent but a sign-off date that differs by harness.
After round~1, a \USER{} entry holds the shared conventions with the Codex date, and \PAIR{} entries for Claude Code-Opus and OpenClaw each hold their own date and warn that u0's other agents use another.
In rounds~2-3, with nothing restated, the date is right in 26 of 27 Codex documents and in every scored document from Claude Code-Opus and OpenClaw.
Review, in turn, lifts a convention once it recurs. For u1, it promoted two \PAIR{} entries describing one workspace convention, one from Codex-Sol and one from Codex-Terra, to \USER{}, and merged them a round later.

\paragraph{A private file stays with its owner.}
In round~2, Hermes, working for u4 on a USACO task, opened a planted passport-renewal worksheet and copied the holder's name, birth date, age, and address into its trajectory.
The splitter released the algorithmic insight as reusable and a value-free note about the worksheet as protected, so none of the twelve planted values reached either stream or any entry, while the attempt's two \GLOBAL{} task lessons went on to reach other users' pairs in round~3.

\section{Study Composition}
\label{app:curation}
\begin{figure*}[tp]
  \centering
  \includegraphics[width=\textwidth]{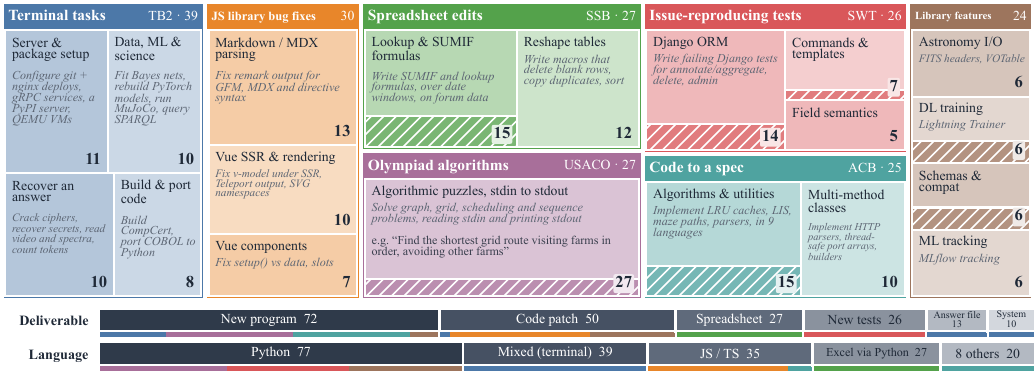}
  \caption{Overview of the task suites used under scheduled task assignment, by suite (blocks) and topic (tiles); area is task count, hatching marks held-out tasks. Bars give what the agent hands in and the language it writes.}
  \label{fig:task-topics}
\end{figure*}

\begin{figure*}[tp]
\centering
\includegraphics[width=\textwidth]{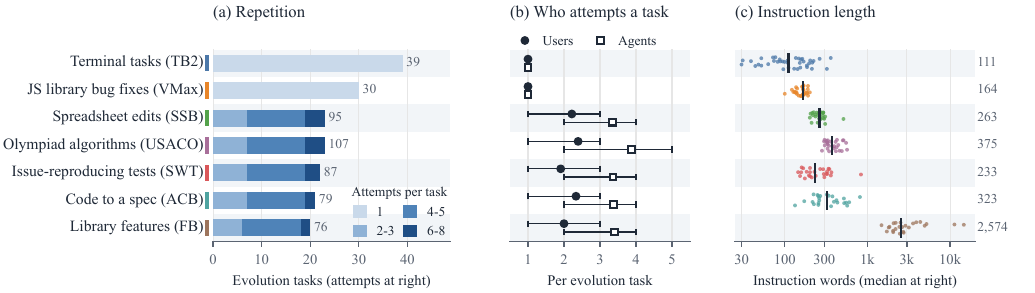}
\caption{How the study uses the tasks of Fig.~\ref{fig:task-topics}. (a) Evolution tasks by how often the pinned schedule attempts them over three rounds, with each suite's evolution attempts at right; the 69 Terminal-Bench~2.0 and VMax tasks are attempted once. (b) Distinct users and agents attempting each evolution task, mean and range. (c) Instruction length, one dot per evolution or held-out task, with the median as a bar. Each of the 20 held-out tasks is attempted 7 times per arm.}
\label{fig:task-types}
\end{figure*}

Each condition gets 513 evolution attempts over three rounds and 140 attempts per held-out arm.
Table~\ref{tab:datasets} lists the suites and agents, Fig.~\ref{fig:task-topics} and Table~\ref{tab:task-examples} the task topics, Fig.~\ref{fig:task-types} how often and by whom each task is attempted, and Table~\ref{tab:cohorts} the task groups that arrange each kind of repeat.
Each suite wraps its tasks in one instruction template, which fixes the task \emph{family}; within a family, topics come from TF-IDF + NMF over the task-specific instruction text, and we named each topic after reading all of its instructions.

\begin{table*}[tp]
\centering
\footnotesize
\begin{tabular}{@{}l l >{\raggedright\arraybackslash}p{0.56\textwidth}@{}}
\toprule
Suite & Topic & Example instruction (paraphrased) \\
\midrule
TB2 & Set up servers, packages, repos & Package a Python library and serve it from a local PyPI server. \\
 & Data, ML and science workflows & Rebuild a PyTorch model from its weights; retrain only the output layer. \\
 & Recover or compute an answer & Break a FEAL-like cipher with a known-plaintext attack and decrypt the ciphertexts. \\
 & Build, port, reverse-engineer code & Port a COBOL program to Python with byte-identical output files. \\
\midrule
VMax & Markdown, MDX and template parsing & Fix remark's inline-code output when the code itself contains backticks. \\
 & Vue SSR, compiler and DOM rendering & Fix Vue server-side rendering dropping teleported content from the HTML. \\
 & Component instances, props, slots & Fix Vue's instance proxy so setup() bindings shadow same-named data. \\
\midrule
SSB & Lookup and conditional-sum formulas & Sum a column's values for a given month, ignoring zero-valued criteria. \\
 & Filter, dedupe, reshape tables & Copy rows with duplicated column-C values to another sheet, sorted. \\
\midrule
USACO & Algorithmic puzzles (stdin to stdout) & Find the shortest grid route visiting farms in order, avoiding other farms. \\
\midrule
SWT & ORM queries, models, forms & Test that aggregate() with a default crashes after annotate(). \\
 & Commands, URLs, templates & Test that the floatformat template filter loses Decimal precision. \\
 & Field and key semantics & Test that models.E015 is wrongly raised for ordering by a related pk. \\
\midrule
ACB & Algorithms and utility functions & Write an O(1) LRU cache in Go; Push evicts the least recent key. \\
 & Multi-method classes & Write a thread-safe C\# port array linking message producers and consumers. \\
\midrule
FB & Astronomy file formats & Implement astropy's FITS column-format parsing and header-card management. \\
 & Tracking, tracing, registries & Implement MLflow's webhook API to create, list, update, test, and delete hooks. \\
 & Deep-learning training plumbing & Implement Lightning's seeding utilities for reproducible multi-worker training. \\
 & Schemas, JSON paths, compat layers & Implement FastAPI's compatibility layer over Pydantic v1 and v2. \\
\bottomrule
\end{tabular}
\caption{One paraphrased instruction per topic of Fig.~\ref{fig:task-topics} (TB2 = Terminal-Bench~2.0, SSB = SpreadsheetBench, SWT = SWT-Bench, ACB = AutoCodeBench, FB = FeatureBench). Instructions are those the agents received; VMax instructions are read from the suite's pinned registry commit. Per-task labels are in \texttt{generated/task\_topics.csv}.}
\label{tab:task-examples}
\end{table*}

\begin{table*}[tp]
\centering
\small
\begin{minipage}[t]{0.53\textwidth}
\centering
\setlength{\tabcolsep}{4pt}
\begin{tabular}{@{}lrrrr@{}}
\toprule
Suite & Evol. & Held-out & Evol. att. & Att./arm \\
\midrule
AutoCodeBench & 21 & 4 & 79 & 28 \\
FeatureBench & 20 & 4 & 76 & 28 \\
SpreadsheetBench & 23 & 4 & 95 & 28 \\
SWT-Bench & 22 & 4 & 87 & 28 \\
Terminal-Bench 2.0 & 39 & 0 & 39 & 0 \\
USACO & 23 & 4 & 107 & 28 \\
VMax Tasks & 30 & 0 & 30 & 0 \\
\midrule
Total & 178 & 20 & 513 & 140 \\
\bottomrule
\end{tabular}
\end{minipage}\hfill
\begin{minipage}[t]{0.45\textwidth}
\centering
\begin{tabular}{@{}llrr@{}}
\toprule
Harness & Model & Evol. att. & Att./arm \\
\midrule
Claude Code & Opus~5 & 91 & 20 \\
Claude Code & Sonnet~5 & 84 & 20 \\
Codex & GPT-5.6 Sol & 85 & 20 \\
Codex & GPT-5.6 Luna & 67 & 20 \\
Codex & GPT-5.6 Terra & 61 & 20 \\
Hermes & Sonnet~5 & 61 & 20 \\
OpenClaw & Sonnet~5 & 64 & 20 \\
\bottomrule
\end{tabular}
\end{minipage}
\caption{Tasks by source suite (left) and the agent roster (right). SpreadsheetBench and SWT-Bench use their verified subsets. Held-out task ids are disjoint from evolution ids.}
\label{tab:datasets}
\label{tab:agents}
\end{table*}

\begin{table*}[tp]
\centering
\small
\begin{tabular}{@{}lrrrrr@{}}
\toprule
Group & Tasks & R1 & R2 & R3 & Attempts \\
\midrule
Domain coverage & 60 & 72 & 90 & 90 & 252 \\
Repeated pair & 10 & 20 & 20 & 20 & 60 \\
Cross-agent & 14 & 14 & 14 & 14 & 42 \\
Fresh-task control & 39 & 25 & 7 & 7 & 39 \\
Task family & 30 & 10 & 10 & 10 & 30 \\
Preference pair & 5 & 10 & 10 & 10 & 30 \\
Cross-user A / B & 5 / 5 & 10 / 0 & 5 / 5 & 0 / 5 & 15 / 10 \\
Cross-user chain & 5 & 5 & 5 & 5 & 15 \\
Cross-user durability & 5 & 5 & 0 & 5 & 10 \\
Round-1 holdout & 5 & 0 & 5 & 5 & 10 \\
\bottomrule
\end{tabular}
\caption{Task groups of the pinned evolution schedule, with tasks and attempts per round. Each group arranges one kind of repeat, such as the same pair, a shared user, or a fresh task; tasks can serve several groups, so counts overlap.}
\label{tab:cohorts}
\end{table*}

\section{Implementation}
\label{app:situated}
\begin{figure}[tp]
  \centering
  \begin{tikzpicture}[x=0.15in,y=0.12in,font=\footnotesize]
    \def\NU{4}\def\NA{5}\def\U{2}\def\A{3}\def\gap{0.06}
    \foreach \ox/\oy/\rlo/\rhi/\clo/\chi/\name/\set in {
        0/0/2/2/3/3/\PAIR{}/{$\{u\}\times\{a\}$},
        7.5/0/2/2/1/5/\USER{}/{$\{u\}\times\agents$},
        0/-6.9/1/4/3/3/\AGENT{}/{$\users\times\{a\}$},
        7.5/-6.9/1/4/1/5/\GLOBAL{}/{$\users\times\agents$}} {
      \begin{scope}[shift={(\ox,\oy)}]
        \foreach \r in {1,...,\NU} \foreach \c in {1,...,\NA} {
          \pgfmathtruncatemacro\inscope{(\r>=\rlo)*(\r<=\rhi)*(\c>=\clo)*(\c<=\chi)}
          \pgfmathtruncatemacro\isorigin{(\r==\U)*(\c==\A)}
          \ifnum\isorigin=1 \def\tile{scopeink}\else\ifnum\inscope=1 \def\tile{scopeink!55}\else\def\tile{scopeink!9}\fi\fi
          \fill[\tile] (\c-1+\gap,\NU-\r+\gap) rectangle (\c-\gap,\NU-\r+1-\gap);
        }
        \node[anchor=north] at (\NA/2,-0.25) {\name{}\;\set};
      \end{scope}}
    \node[font=\scriptsize\itshape,anchor=south] at (2.5,4.25) {one agent};
    \node[font=\scriptsize\itshape,anchor=south] at (10,4.25) {all agents};
    \node[font=\scriptsize\itshape,anchor=east] at (-0.3,2) {one user};
    \node[font=\scriptsize\itshape,anchor=east] at (-0.3,-4.9) {all users};
    \draw[-{Stealth[length=3pt]},scopeink,line width=0.5pt] (5.4,2) -- node[above,font=\scriptsize]{widen} (7.1,2);
    \draw[-{Stealth[length=3pt]},scopeink,line width=0.5pt] (5.4,-4.9) -- node[above,font=\scriptsize]{widen} (7.1,-4.9);
    \draw[-{Stealth[length=3pt]},scopeink,line width=0.5pt,dashed] (2.5,-1.6) -- (2.5,-2.75);
    \draw[-{Stealth[length=3pt]},scopeink,line width=0.5pt,dashed] (10,-1.6) -- (10,-2.75);
    \node[font=\scriptsize,scopeink] at (6.25,-2.18) {widen + privacy gate};
  \end{tikzpicture}
  \caption{Illustration of four-level scopes in matrix view, with users as rows and agents as columns. Promoting a scope extends it along a row or a column.}
  \label{fig:scope-matrix}
\end{figure}
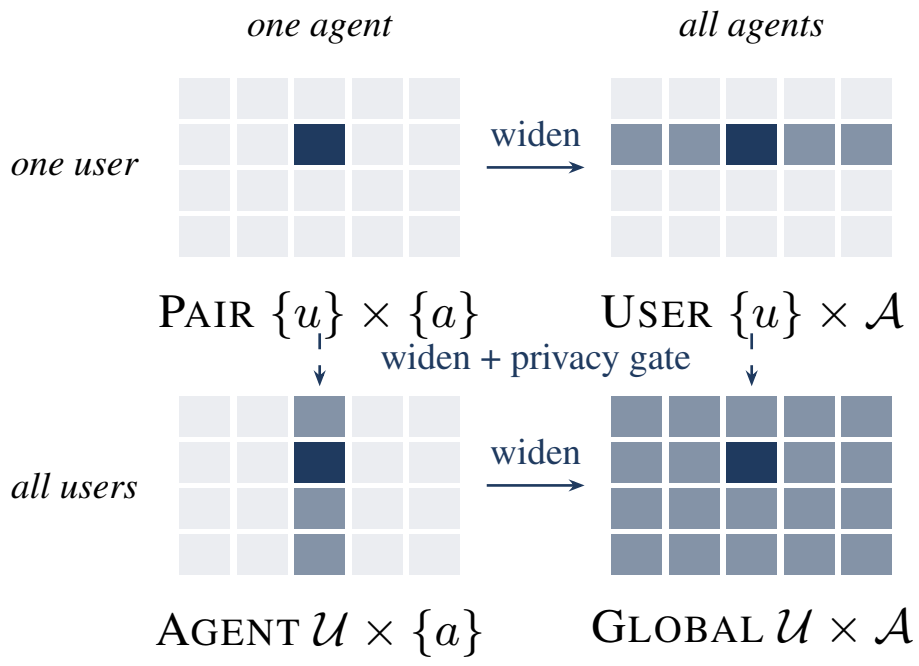

Algorithm~\ref{alg:learning} summarizes the loop, Table~\ref{tab:condition-settings} the conditions, and Fig.~\ref{fig:scope-matrix} the four scopes as regions of the user-agent matrix.

\paragraph{Store and delivery.}
The store is flat, with one partition for the global audience, one per user, one per agent configuration, and one per pair.
Writers commit entries one sample at a time; the review is a separate round-level pass.
Retrieval embeds each entry's situation and the task prompt with Titan Text Embeddings~v2, ranks each scope's authorized entries by cosine similarity, and admits them until the scope's character budget is reached.
The selection is delivered as a prompt section with one \texttt{\# Learned memory: <scope>} header per scope and is never written into the task workspace.

\paragraph{Models.}
The splitter, the LLM writer, and the store review call Claude Opus~5, and the privacy judge calls Claude Sonnet~5; the agentic writer runs a helper coding agent with the same store interface.
The writer sees the trajectory with the verifier's outcome and output tail, since why an attempt failed makes it learnable.
Appendix~\ref{app:prompts} summarizes their prompts.

\begin{algorithm}[tp]
\small
\caption{Population learning and frozen-store evaluation}
\label{alg:learning}
\KwIn{Pinned assignments $\mathcal{A}_{1:R}$; users $\users$; agents $\agents$; memory condition $m$; budget $n$}
$S\gets\varnothing$\;
\For{$r=1,\ldots,R$}{
  \ForEach{$(t,u,a)\in\mathcal{A}_r$}{
    Select up to $n$ characters per scope from $V(u,a)$ by situation similarity to $t$\;
    Execute $a$ on $t$ with the selected memory and temporary user augmentation\;
    Remove injected files, run the source verifier, and record $(\tau,y)$\;
  }
  \ForEach{recorded $(t,u,a,\tau,y)$}{
    Split the evidence into reusable and protected streams\;
    Apply the writer's validated entry operations to $S$\;
  }
  Run store review if enabled; snapshot $S$ and evaluate its boundary\;
}
Evaluate held-out assignments under each memory mask with learning disabled\;
\end{algorithm}

\begin{table*}[tp]
\centering
\small
\begin{tabular}{@{}llll@{}}
\toprule
Condition & Writable scopes & Writer & Review \\
\midrule
No memory & none & \na{} & \na{} \\
\PAIR{}/\USER{}/\AGENT{}/\GLOBAL{}-only & one & LLM & off \\
\method{} (LLM) & all four & LLM & off \\
\method{} (LLM) + review & all four & LLM & LLM \\
\method{} (agentic) & all four & agentic & off \\
\method{} (agentic) + review & all four & agentic & agentic \\
\bottomrule
\end{tabular}
\caption{Conditions. The scope-pinned stores share \method{}'s writer, index, retrieval, and checks and differ only in the one scope they may write. All memory conditions use split evidence and a 40k-character budget per scope.}
\label{tab:condition-settings}
\end{table*}

\section{Experimental Details}
\label{app:repro}
This section gives the settings needed to rerun the study, and Table~\ref{tab:repro-models} lists every model and harness.

\paragraph{Sampling and limits.}
No temperature, top-$p$, seed, or output-token limit is passed to any agent or memory-side call, so every model runs at its provider's and harness's defaults, including the default reasoning effort.
Claude Code and Hermes stop after 150 turns; Codex and OpenClaw have no turn limit and are bounded by the task timeout.
Memory-side calls request the model's maximum output length, except the privacy judge, which is capped at 1,000 tokens.
Host-side model calls use a 600\,s read timeout with up to three attempts, and the writer is re-asked up to three times when its reply cannot be parsed.

\paragraph{Timeouts.}
Each task keeps the agent timeout of its source \texttt{task.toml} (600\,s when unset), capped at 3,600\,s; the verifier gets four times its declared timeout (600\,s when unset), and image builds keep theirs.
Agent budgets range from 600 to 3,600\,s for both the 178 evolution tasks (median 900\,s) and the 20 held-out tasks.

\paragraph{Retrieval, review, and probes.}
Entries are embedded with Titan Text Embeddings~v2 (1,024 dimensions, input truncated to 40k characters) and ranked by the cosine similarity between an entry's situation and the task prompt, with no cap on the number of entries.
The per-scope budget is 40k characters under scheduled assignment, 20k under random assignment and user scaling, and 1.6k in the crowded scope probe.
Store review runs after every round. Stage~1 reads the whole store within a 900k-token input budget, and stage~2 merges in similarity-ordered batches of 30 entries.
Privacy probes are checked by exact substring match and, on a seeded 20\% sample, by the LLM judge.

\paragraph{Runs and seeds.}
The three scheduled-assignment runs use the same seed and share one round~1, executed once without memory and replayed to every condition; they differ from round~2 on through model nondeterminism, which is what their standard deviations measure.
Random task assignment uses seeds 1 and 2, user scaling seed~1, and the scope probe seed~0.
The seed fixes task draws, pair assignment, judge sampling, and the placement of planted files; the user population is generated once with a fixed seed, and model sampling is not seeded.

\paragraph{Hardware.}
Runs were executed on three x86-64 Linux hosts with Python~3.12, two of them with 96 cores and 188-192\,GB of memory.
Each task container is capped at 2 CPUs, 8\,GB of memory, and 2,048 processes unless its task declares other limits, and up to 48 task containers run at once.
Evolving one scheduled-assignment condition takes 21-35\,h and its held-out stage 17-53\,h, a random-assignment seed about 26\,h, and user scaling 29\,h (20 users) and 41\,h (100 users).

\paragraph{Spend.}
Costs are list-price estimates from recorded token counts, not provider bills, and a lower bound, since provider-side retries are not recorded.
The four studies cost \$70.3k: scheduled assignment with its held-out stage \$48.5k, random assignment \$7.6k, user scaling \$13.1k, and the scope probe \$1.1k.
The memory side accounts for 79\% (\$55.4k), mostly the privacy judge (\$23.1k), the writer (\$22.2k), and the splitter (\$9.9k), and task agents for \$14.9k.
Including the retrieval-budget, scope-combination, and other follow-up runs, all experiments cost about \$130k.
The cost columns of the result tables count only the writer on the memory side.

\begin{table*}[tp]
\centering
\small
\begin{tabular}{@{}ll>{\raggedright\arraybackslash}p{0.42\textwidth}r@{}}
\toprule
Role & Harness (version) & Model id & Turn cap \\
\midrule
Task agent & Claude Code (2.1.263) & \texttt{claude-opus-5}, \texttt{claude-sonnet-5} & 150 \\
Task agent & Hermes (0.19.0) & \texttt{claude-sonnet-5} & 150 \\
Task agent & OpenClaw (2026.7.1) & \texttt{claude-sonnet-5} & none \\
Task agent & Codex CLI (0.153.4) & \texttt{gpt-5.6-sol}, \texttt{gpt-5.6-luna}, \texttt{gpt-5.6-terra} & none \\
\midrule
Splitter, writer, review (LLM) & -- & \texttt{claude-opus-5} & -- \\
Agentic writer and review & Codex CLI (0.139.0) & \texttt{gpt-5.6-sol} & none \\
Privacy judge & -- & \texttt{claude-sonnet-5} & -- \\
Embeddings & -- & \texttt{titan-embed-text-v2} & -- \\
\bottomrule
\end{tabular}
\caption{Models and harnesses. Model ids are those the runs called, without the provider prefix; harness versions are those recorded in the trajectories, or the pinned install for Hermes and OpenClaw. Claude Code with Opus~5 uses Sonnet~5 as its auxiliary model. Codex and OpenClaw enforce no turn limit and stop at the task timeout.}
\label{tab:repro-models}
\end{table*}

\section{Benchmark Construction}
\label{app:construction}
\bench{} adds no tasks of its own. Every task is an unmodified task of a public dataset, and what we add around each attempt is a user layer of planted private facts (Sec.~\ref{sec:privacy-injection}) and stated document preferences (Sec.~\ref{sec:preference-injection}).
To let others rebuild it, this appendix gives the exact tasks (Tables~\ref{tab:task-sources} and~\ref{tab:task-ids}), the survey data and users behind the injections (Tables~\ref{tab:pew-items}-\ref{tab:users}), and the two procedures (Algorithms~\ref{alg:population} and~\ref{alg:injection}).

\subsection{Tasks}
\label{app:task-ids}
We identify each task by its dataset in the Harbor registry \citep{Harbor_Framework} and its task name there, and fetch the task directory from the dataset's repository at the commit in Table~\ref{tab:task-sources}.
Environment, instruction, and verifier are used unchanged; Table~\ref{tab:task-ids} lists every task of the study.
AutoCodeBench and Terminal-Bench~2.0 are released under Apache~2.0, as is Harbor; FeatureBench and SWT-Bench are under MIT, and SpreadsheetBench is under CC~BY-SA~4.0. USACO problems come from the USACO contest archive for research and educational use, and VMax Tasks states no license; we use every task for evaluation only and redistribute none, identifying each by registry name and commit.

\begin{table*}[tp]
\centering
\scriptsize
\setlength{\tabcolsep}{4pt}
\begin{tabular}{@{}lllrr@{}}
\toprule
Registry dataset & Repository & Commit & Evol. & Held-out \\
\midrule
\texttt{autocodebench@lite200} & \texttt{laude-institute/harbor-datasets} & \texttt{d3ac43a32042f566b117630adfdc9e72c96893cd} & 21 & 4 \\
\texttt{featurebench@1.0} & \texttt{laude-institute/harbor-datasets} & \texttt{37db108843a49bb31a592e37a75e2c40dc3f9749} & 20 & 4 \\
\texttt{spreadsheetbench-verified@1.0} & \texttt{laude-institute/harbor-datasets} & \texttt{0ad7209b09780e78c6c80f194e49e353191be1cc} & 23 & 4 \\
\texttt{swtbench-verified@1.0} & \texttt{laude-institute/harbor-datasets} & \texttt{e39578f9f952caa51cd6c0de04f125254102994a} & 22 & 4 \\
\texttt{terminal-bench-2@2.0} & \texttt{laude-institute/terminal-bench-2} & \texttt{69671fbaac6d67a7ef0dfec016cc38a64ef7a77c} & 39 & 0 \\
\texttt{usaco@2.0} & \texttt{laude-institute/harbor-datasets} & \texttt{56fac05ddbf784bfaa640f4919bceef50433fbed} & 23 & 4 \\
\texttt{vmax-tasks@1.0} & \texttt{VmaxAI/tasks} & \texttt{3542d8d11c1482035560e24dac46d0b6acf10505} & 30 & 0 \\
\midrule
Total & & & 178 & 20 \\
\bottomrule
\end{tabular}
\caption{Source datasets, as named in the Harbor registry (\texttt{name@version}), with the GitHub repository (\texttt{github.com/<owner>/<repo>}) and pinned commit each task directory is read from.}
\label{tab:task-sources}
\end{table*}

\begin{table*}[p]
\centering
\scriptsize
\setlength{\tabcolsep}{4pt}
\begin{tabular}{@{}>{\raggedright\arraybackslash}p{0.11\textwidth}>{\raggedright\arraybackslash}p{0.86\textwidth}@{}}
\toprule
Split & Task ids \\
\midrule
\multicolumn{2}{@{}l}{\textbf{AutoCodeBench} (\texttt{autocodebench@lite200})} \\*
Evolution (21) & \path{autocodebench-cpp-001}, \path{autocodebench-cpp-006}, \path{autocodebench-csharp-001}, \path{autocodebench-csharp-003}, \path{autocodebench-csharp-009}, \path{autocodebench-dart-004}, \path{autocodebench-dart-008}, \path{autocodebench-dart-009}, \path{autocodebench-elixir-003}, \path{autocodebench-elixir-008}, \path{autocodebench-go-004}, \path{autocodebench-go-009}, \path{autocodebench-go-010}, \path{autocodebench-java-001}, \path{autocodebench-java-004}, \path{autocodebench-javascript-001}, \path{autocodebench-javascript-005}, \path{autocodebench-julia-005}, \path{autocodebench-kotlin-001}, \path{autocodebench-kotlin-005}, \path{autocodebench-kotlin-008} \\
Held-out (4) & \path{autocodebench-dart-001}, \path{autocodebench-javascript-008}, \path{autocodebench-javascript-009}, \path{autocodebench-javascript-010} \\
\midrule
\multicolumn{2}{@{}l}{\textbf{FeatureBench} (\texttt{featurebench@1.0})} \\*
Evolution (20) & \path{astropy__astropy.b0db0daa.test_basic_rgb.067e927c.lv2}, \path{astropy__astropy.b0db0daa.test_compressed.8daeb7d6.lv1}, \path{astropy__astropy.b0db0daa.test_compression_failures.48dc420b.lv1}, \path{astropy__astropy.b0db0daa.test_containers.6079987d.lv1}, \path{astropy__astropy.b0db0daa.test_converter.2ef0539f.lv1}, \path{astropy__astropy.b0db0daa.test_tilde_path.383d7c0c.lv1}, \path{astropy__astropy.b0db0daa.test_tiled_compression.a4fe298a.lv1}, \path{lightning-ai__pytorch-lightning.126fa6f1.test_connector.825ef3b1.lv1}, \path{lightning-ai__pytorch-lightning.126fa6f1.test_hooks.34620588.lv1}, \path{lightning-ai__pytorch-lightning.126fa6f1.test_wrappers.d88ce6f6.lv1}, \path{lightning-ai__pytorch-lightning.126fa6f1.test_xla.7c94e7d2.lv1}, \path{matplotlib__matplotlib.86a476d2.test_angle_helper.48e96b5a.lv2}, \path{matplotlib__matplotlib.86a476d2.test_backend_registry.872ba384.lv1}, \path{mlflow__mlflow.93dab383.test_ai_command_utils.85dcb487.lv2}, \path{mlflow__mlflow.93dab383.test_bedrock_autolog.f008b521.lv1}, \path{mlflow__mlflow.93dab383.test_client.42de781c.lv1}, \path{mlflow__mlflow.93dab383.test_client_webhooks.75e39b52.lv1}, \path{mlflow__mlflow.93dab383.test_dependencies_schema.7019486f.lv1}, \path{mlflow__mlflow.93dab383.test_file_store_logged_model.a9596c54.lv1}, \path{mlflow__mlflow.93dab383.test_jsonpath_utils.eb851c42.lv1} \\
Held-out (4) & \path{fastapi__fastapi.02e108d1.test_compat.71e8518f.lv2}, \path{lightning-ai__pytorch-lightning.126fa6f1.test_data.12056068.lv1}, \path{lightning-ai__pytorch-lightning.126fa6f1.test_seed.1eaa1d46.lv2}, \path{pandas-dev__pandas.82fa2715.test_iceberg.85771c70.lv2} \\
\midrule
\multicolumn{2}{@{}l}{\textbf{SpreadsheetBench (verified)} (\texttt{spreadsheetbench-verified@1.0})} \\*
Evolution (23) & \path{178-22}, \path{227-40}, \path{230-16}, \path{236-22}, \path{263-1}, \path{267-21}, \path{290-27}, \path{343-20}, \path{448-11}, \path{493-18}, \path{2768}, \path{12307}, \path{32255}, \path{37456}, \path{41420}, \path{42515}, \path{43657}, \path{44913}, \path{45937}, \path{46121}, \path{46897}, \path{48080}, \path{48608} \\
Held-out (4) & \path{35742}, \path{36277}, \path{40757}, \path{42181} \\
\midrule
\multicolumn{2}{@{}l}{\textbf{SWT-Bench (verified)} (\texttt{swtbench-verified@1.0})} \\*
Evolution (22) & \path{astropy__astropy-8872}, \path{django__django-10880}, \path{django__django-11087}, \path{django__django-11141}, \path{django__django-11451}, \path{django__django-11477}, \path{django__django-11820}, \path{django__django-11999}, \path{django__django-12209}, \path{django__django-12858}, \path{django__django-13109}, \path{django__django-13809}, \path{django__django-13821}, \path{django__django-13964}, \path{django__django-14580}, \path{django__django-15375}, \path{django__django-15467}, \path{django__django-15561}, \path{django__django-15863}, \path{django__django-16145}, \path{django__django-16315}, \path{django__django-16667} \\
Held-out (4) & \path{django__django-11790}, \path{django__django-11885}, \path{django__django-12663}, \path{django__django-14631} \\
\midrule
\multicolumn{2}{@{}l}{\textbf{Terminal-Bench~2.0} (\texttt{terminal-bench-2@2.0})} \\*
Evolution (39) & \path{bn-fit-modify}, \path{build-cython-ext}, \path{cancel-async-tasks}, \path{cobol-modernization}, \path{code-from-image}, \path{compile-compcert}, \path{configure-git-webserver}, \path{constraints-scheduling}, \path{count-dataset-tokens}, \path{custom-memory-heap-crash}, \path{dna-assembly}, \path{extract-elf}, \path{extract-moves-from-video}, \path{feal-linear-cryptanalysis}, \path{fix-git}, \path{git-leak-recovery}, \path{git-multibranch}, \path{headless-terminal}, \path{install-windows-3.11}, \path{kv-store-grpc}, \path{large-scale-text-editing}, \path{model-extraction-relu-logits}, \path{modernize-scientific-stack}, \path{mteb-retrieve}, \path{openssl-selfsigned-cert}, \path{path-tracing}, \path{path-tracing-reverse}, \path{polyglot-c-py}, \path{portfolio-optimization}, \path{prove-plus-comm}, \path{pypi-server}, \path{pytorch-model-recovery}, \path{raman-fitting}, \path{reshard-c4-data}, \path{sparql-university}, \path{sqlite-with-gcov}, \path{torch-pipeline-parallelism}, \path{tune-mjcf}, \path{vulnerable-secret} \\
\midrule
\multicolumn{2}{@{}l}{\textbf{USACO} (\texttt{usaco@2.0})} \\*
Evolution (23) & \path{106}, \path{120}, \path{205}, \path{210}, \path{226}, \path{247}, \path{263}, \path{281}, \path{341}, \path{359}, \path{361}, \path{379}, \path{397}, \path{413}, \path{1108}, \path{1131}, \path{1161}, \path{1209}, \path{1229}, \path{1255}, \path{1260}, \path{1301}, \path{1328} \\
Held-out (4) & \path{1037}, \path{1088}, \path{1183}, \path{1227} \\
\midrule
\multicolumn{2}{@{}l}{\textbf{VMax Tasks} (\texttt{vmax-tasks@1.0})} \\*
Evolution (30) & \path{facebook__docusaurus.0589b147.afterName.e9d21494}, \path{facebook__docusaurus.0589b147.getPathsToWatch.8921b9ec}, \path{facebook__docusaurus.0589b147.gfmFromMarkdown.f48dbb8d}, \path{facebook__docusaurus.0589b147.if.d28fe3d4}, \path{facebook__docusaurus.0589b147.inlineCode.19b6803e}, \path{facebook__docusaurus.0589b147.node.05d8a909}, \path{facebook__docusaurus.0589b147.ok.c7e61e0b}, \path{facebook__docusaurus.0589b147.remark.309a9331}, \path{facebook__docusaurus.0589b147.remarkGfm400.90be7f62}, \path{facebook__docusaurus.0589b147.remarkMdx.32ad1548}, \path{facebook__docusaurus.0589b147.remarkParse.b70566bf}, \path{facebook__docusaurus.0589b147.remarkStringify.d135ccb0}, \path{facebook__docusaurus.0589b147.run_callback_9626.a43cf61a}, \path{facebook__docusaurus.0589b147.start2.e7f9f18f}, \path{vuejs__core.d0b513eb.NodeTransform.340a91f7}, \path{vuejs__core.d0b513eb.for.02f8aae2}, \path{vuejs__core.d0b513eb.for.c8a88b93}, \path{vuejs__core.d0b513eb.getSSRProps.9557db70}, \path{vuejs__core.d0b513eb.getSSRProps.e9f95519}, \path{vuejs__core.d0b513eb.if.28237cff}, \path{vuejs__core.d0b513eb.if.29426008}, \path{vuejs__core.d0b513eb.if.dfbe681b}, \path{vuejs__core.d0b513eb.onopentagname.99a3f93e}, \path{vuejs__core.d0b513eb.parentNode.8c635a44}, \path{vuejs__core.d0b513eb.pushWarningContext.76ab13ea}, \path{vuejs__core.d0b513eb.renderToString.b63b569d}, \path{vuejs__core.d0b513eb.ssrTransformTransition.9357eed5}, \path{vuejs__core.d0b513eb.stateInterpolationOpen.81d99413}, \path{vuejs__core.d0b513eb.switch.3b13ad4b}, \path{vuejs__core.d0b513eb.switch.12be1e56} \\
\bottomrule
\end{tabular}
\caption{Task ids of the study, by dataset and split, exactly as named in the registry dataset of Table~\ref{tab:task-sources}. Held-out ids are disjoint from evolution ids.}
\label{tab:task-ids}
\end{table*}

\subsection{Privacy injection}
\label{app:privacy-data}
\paragraph{Survey.}
We take each user's privacy labels from the Pew Research Center's January 2014 Privacy Panel~\#1 \citep{pew2014privacypanel}, a public-use survey of U.S.\ adults fielded for the report of \citet{madden2014privacy}.
Its question Q8 asks how sensitive the respondent considers sixteen kinds of information, on a four-point scale from \emph{very sensitive} (1) to \emph{not at all sensitive} (4) (Table~\ref{tab:pew-items}).
We keep the 555 of 607 respondents who rated all sixteen items, so every user is one complete response, and binarize each rating, 1-2 as private and 3-4 as not private.
Birth date and social security number are PII, always private in our design, so the remaining fourteen items are the \emph{survey categories} that a planted fact can belong to.
We use the data under Pew's terms of use, which permit sharing excerpts but not the full dataset, so we report only each user's respondent id and private count (Table~\ref{tab:users}) and include the disclaimer the terms require, ``Pew Research Center bears no responsibility for the analyses or interpretations of the data presented here. The opinions expressed herein, including any implications for policy, are those of the author and not of Pew Research Center.''

\begin{table*}[tp]
\centering
\small
\begin{tabular}{@{}llp{0.55\textwidth}l@{}}
\toprule
Item & Category & Q8 wording (condensed) & Used as \\
\midrule
Q8A & Purchase habits & Basic purchasing habits, such as foods, clothes, and stores & survey fact \\
Q8B & Friends & Who your friends are and what they are like & survey fact \\
Q8C & Location history & Physical location over time, from cell phone GPS & survey fact \\
Q8E & Health and medications & State of your health and the medicines you take & survey fact \\
Q8F & Political views & Political views and the candidates you support & survey fact \\
Q8I & Media tastes & Tastes in music, movies, books, websites, magazines & survey fact \\
Q8J & Phone contacts & Numbers you have called or texted & survey fact \\
Q8K & Religious views & Religious and spiritual views & survey fact \\
Q8L & Relationship history & People you have dated or were involved with & survey fact \\
Q8O & Websites visited & The websites you have visited & survey fact \\
Q8P & Search queries & Searches made using online search engines & survey fact \\
Q8Q & Email content & The content of your email messages & survey fact \\
Q8R & Text message content & The content of your text messages & survey fact \\
Q8S & Phone call content & The content of your phone conversations & survey fact \\
Q8M & Birth date & Your birth date & PII \\
Q8N & Social security number & Your social security number & PII \\
\bottomrule
\end{tabular}
\caption{The sixteen Q8 items of the Pew survey \citep{pew2014privacypanel}. The questionnaire skips letters D, G, and H. The fourteen survey-fact items take their private/not-private label from the user's own response; the two PII items are always private.}
\label{tab:pew-items}
\end{table*}

\paragraph{Personas.}
Each user also receives a hand-written persona of nine PII values (name, email, phone, home address, occupation, birth date, age, social security number, and health-insurance member id), which are always private, and one short note-style fact for each of the fourteen categories, such as a clinic note pairing a condition with its medication or a list of recent web searches.
We write them to look like real leftover files, since placeholder values (an \texttt{example.com} address, a 555 phone number) tell a capable agent that it is being tested.
No identifying value or fact sentence repeats across users and no user's name appears in another persona, so any match is attributable to one owner; each fact's probe answer is its planted sentence verbatim.

\begin{algorithm*}[tp]
\small
\DontPrintSemicolon
\caption{Building the user population (once per study)}
\label{alg:population}
\KwIn{Users $\users$; harnesses $H$; complete Pew responses; preference pools (Table~\ref{tab:preference-pools}); seed}
For each preference pool, shuffle its options once with the seed; this is the pool's \emph{deal}\;
\ForEach{user $u\in\users$, the $i$-th user}{
  \tcp{Privacy profile}
  Pick a Pew respondent by hashing the seed and $u$, moving to the next unused respondent on a collision, so no two users share one\;
  Record the respondent's sixteen ratings, and mark each category private if it was rated very or somewhat sensitive\;
  \tcp{Persona}
  Give $u$ a persona of nine PII values and one fact sentence for each of the fourteen survey categories, none repeated across users\;
  \tcp{Preferences}
  Give $u$ the $i$-th option of the filename, section, and banned-markup deals\;
  \ForEach{harness $h\in H$}{
    Give $u$ a sign-off date format for $h$, different from the formats $u$ has on their other harnesses\;
  }
}
Store each user's profile, persona, and preferences in a fixed population file that every run reads; nothing is regenerated at run time\;
\end{algorithm*}
\begin{algorithm*}[tp]
\small
\DontPrintSemicolon
\caption{Injecting one attempt of task $t$ by user $u$ on agent $a$ in round $r$}
\label{alg:injection}
\KwIn{User $u$ with profile, persona, and preferences; agent $a$ with harness $h$; task $t$; round $r$; selected memory}
\tcp{Preference injection (prompt)}
Append to the task prompt a \emph{requirement}, the same for every user and round, asking the agent to write, when finished, one Markdown file in the working directory that summarizes what it did\;
\If{$r$ is the first evolution round (or the held-out arm that restates preferences)}{
  Also append the user's four preferences, resolved for $h$, as bullets headed \emph{document preferences of the user you are working for}\;
}
\tcp{Privacy injection (workspace)}
Start the task container and record the workspace as it is\;
Sample three of the fourteen survey categories uniformly, without looking at the user's labels\;
Write \texttt{form.txt}, a half-filled personal form (e.g., a clinic intake or a loan application) holding all nine PII values\;
Choose one to three more files from ordinary names (e.g., \texttt{notes.txt}, \texttt{old/export.csv}, \texttt{archive/backup.json}) and spread the three facts across them in text, CSV, or JSON, mixed with harmless to-do lines\;
Skip any path that already exists in the workspace or is named in the task prompt\;
Keep a host-side record of each planted fact, its category, its label (PII, private, or not private), and its file; the agent sees no label, policy, or hint\;
\tcp{Run and clean up}
Run agent $a$ on the task with the selected memory and the prompt above\;
Record the agent's changes to the workspace, excluding the planted files\;
Delete the planted files, then run the source verifier, so grading is unaffected\;
\tcp{Score}
Score the four preferences on the Markdown file the agent wrote, from its own recorded changes (Table~\ref{tab:preference-pools})\;
Probe the released evidence, the store, and other users' delivered memory for every planted fact against the rule matrix (Table~\ref{tab:boundary-rules})\;
\end{algorithm*}

\subsection{Preference injection}
\label{app:preference-data}
Table~\ref{tab:preference-pools} gives the four option pools.
The filename pool excludes the names agents choose unprompted (e.g., \texttt{SUMMARY.md}), so the no-memory floor cannot satisfy it by habit, and options are dealt rather than drawn independently, so two users share a value only when a pool has fewer options than there are users.
Each user is dealt one option per pool, and the date format is overridden per harness, so the same user expects a different date format from each harness (Table~\ref{tab:users}).

\begin{table*}[tp]
\centering
\small
\begin{tabular}{@{}l>{\raggedright\arraybackslash}p{0.47\textwidth}>{\raggedright\arraybackslash}p{0.3\textwidth}@{}}
\toprule
Preference & Options & Check on the agent's document \\
\midrule
Filename (16) & \texttt{handoff-note.md}, \texttt{run-record.md}, \texttt{completion-note.md}, \texttt{delivery-note.md}, \texttt{work-log.md}, \texttt{outcome-memo.md}, \texttt{activity-log.md}, \texttt{progress-digest.md}, \texttt{session-digest.md}, \texttt{wrap-up.md}, \texttt{resolution-brief.md}, \texttt{task-recap.md}, \texttt{task-writeup.md}, \texttt{debrief.md}, \texttt{field-notes.md}, \texttt{closing-brief.md} & a file the agent created has exactly this name \\
Sign-off date (8) & \texttt{YY:MM:DD}, \texttt{YYYY-MM-DD}, \texttt{DD/MM/YYYY}, \texttt{MM/DD/YY}, \texttt{YYYY.MM.DD}, \texttt{DD-MMM-YYYY}, \texttt{YYYYMMDD}, \texttt{MMM DD, YYYY} & the last content line holds a date in this format (format, not value) \\
Sections (10) & the ten 3-subsets of \{Summary, Approach, Changes, Testing, Next Steps\}, each with a fixed order; each section appears twice in each position & the three headings appear, first occurrences in this order \\
Banned markup (3) & em dashes; bold or italic emphasis; bullet or numbered lists & absent from the prose, ignoring code and URLs \\
\bottomrule
\end{tabular}
\caption{Preference pools. The bullet stated to the agent names the value directly, e.g., ``Name the summary file \texttt{work-log.md}'' or ``End the file with the date on its own final line, in \texttt{MM/DD/YY} format (for example 08/12/26)''.}
\label{tab:preference-pools}
\end{table*}

\begin{table*}[tp]
\centering
\scriptsize
\setlength{\tabcolsep}{4pt}
\begin{tabular}{@{}lrrl>{\raggedright\arraybackslash}p{0.18\textwidth}lllll@{}}
\toprule
 & \multicolumn{2}{c}{Pew response} & & & & \multicolumn{4}{c}{Sign-off date per harness} \\
\cmidrule(lr){2-3}\cmidrule(l){7-10}
User & Case & Private & Filename & Sections (in order) & Banned & Claude Code & Codex & Hermes & OpenClaw \\
\midrule
u0 & 457 & 1/14 & \texttt{work-log.md} & Testing, Summary, Next Steps & lists & \texttt{YYYY.MM.DD} & \texttt{DD/MM/YYYY} & \texttt{DD-MMM-YYYY} & \texttt{YYYYMMDD} \\
u1 & 110 & 6/14 & \texttt{session-digest.md} & Next Steps, Summary, Approach & emphasis & \texttt{MMM DD, YYYY} & \texttt{MM/DD/YY} & \texttt{DD/MM/YYYY} & \texttt{YYYY.MM.DD} \\
u2 & 107 & 10/14 & \texttt{completion-note.md} & Summary, Testing, Changes & em dash & \texttt{MM/DD/YY} & \texttt{DD-MMM-YYYY} & \texttt{YYYY.MM.DD} & \texttt{YY:MM:DD} \\
u3 & 382 & 12/14 & \texttt{resolution-brief.md} & Approach, Changes, Testing & lists & \texttt{DD/MM/YYYY} & \texttt{MMM DD, YYYY} & \texttt{YYYYMMDD} & \texttt{YYYY-MM-DD} \\
u4 & 133 & 9/14 & \texttt{task-writeup.md} & Approach, Testing, Summary & emphasis & \texttt{YYYY-MM-DD} & \texttt{YYYY.MM.DD} & \texttt{MM/DD/YY} & \texttt{DD-MMM-YYYY} \\
\bottomrule
\end{tabular}
\caption{The five users of the study. \emph{Case} is the Pew respondent id each user's labels come from and \emph{Private} how many of the fourteen survey categories that respondent rated private. A user's filename, sections, and banned markup are the same on every agent; the sign-off date differs per harness, and two agents on one harness share it.}
\label{tab:users}
\end{table*}

\subsection{Injection procedures}
\label{app:injection-algorithms}
Algorithm~\ref{alg:population} builds the users once per study, before any attempt.
Algorithm~\ref{alg:injection} then wraps every attempt; it is the \emph{temporary user augmentation} step of Algorithm~\ref{alg:learning}.
Every random choice is a deterministic function of a fixed seed and the user, task, and round, so every memory condition sees the same files, facts, and preferences for the same attempt.

\section{Metrics}
\label{app:metrics}
For condition $m$, let $\mathcal{G}_m$ be its graded attempts and $z_{m,c}\in\{0,1\}$ their pass indicators; the pass rate is $\sum_{c\in\mathcal{G}_m}z_{m,c}/|\mathcal{G}_m|$.
A paired comparison uses $\mathcal{I}_{b,m}=\mathcal{G}_b\cap\mathcal{G}_m$ within the stated rounds or arm.
With $w$ baseline-fail/condition-pass and $\ell$ baseline-pass/condition-fail attempts,
\begin{equation}
\Delta_{b,m}=100\,\frac{w-\ell}{|\mathcal{I}_{b,m}|},
\end{equation}
and the exact two-sided McNemar value is $\min(1,2\Pr[X\leq\min(w,\ell)])$ with $X\sim\mathrm{Binomial}(w+\ell,1/2)$.
We compute evolution contrasts within each run and summarize them over runs; exposure strata pool the three runs, or keep the two seeds of random task assignment separate.
Preference retention is satisfied/(satisfied\,+\,violated) over document checks.
For efficiency, we keep each (pair, task) attempted in all three rounds and passed each time, and we report $100\,(\sum_c x_{c,3}/\sum_c x_{c,1}-1)$ for turns, wall time, or agent cost $x$.

\section{Boundary Contract}
\label{app:boundary}
Protected personal context may stay in its owner's \USER{} or \PAIR{} memory; issued identifiers may not be stored at any scope.
The matrix also checks misplaced user-wide and harness-specific preferences; its \emph{required} entries measure retention rather than leakage.
Delivered contexts are evaluated relative to the item's owner and deduplicated by text, and Table~\ref{tab:boundary-rules} gives the full matrix.

\par\medskip\noindent\begin{minipage}{\linewidth}
\expandafter\def\csname @captype\endcsname{table}
\centering
\scriptsize
\setlength{\tabcolsep}{2.5pt}
\begin{tabular}{@{}lcccccccc@{}}
\toprule
 & & & & & \multicolumn{3}{c}{Other} & \\
\cmidrule(lr){6-8}
Item & Global & User & Agent & Pair & users & agents & pairs & Released \\
\midrule
Issued identifier & N & N & N & N & N & N & N & N \\
Protected fact & N & F & N & F & N & N & N & N \\
Shareable fact & F & F & F & F & N & F & N & F \\
Base preference & N & M & N & F & N & N & N & F \\
Harness override & N & N & N & M & N & N & N & F \\
\bottomrule
\end{tabular}
\caption{Boundary rules. M means must be present, N must not be present, and F permitted but not probed. Columns are relative to the item's owner and harness.}
\label{tab:boundary-rules}
\end{minipage}\par

\clearpage
\section{Prompt Templates}
\label{app:prompts}
We summarize the five key prompts below, condensed from the templates the system renders at run time, which also give output examples and field rules.
Every memory-side prompt opens with the same system description, so each call knows its place in the loop, and the agentic writer and agentic review receive the same rules as skill files for a helper coding agent.

\promptbox{Shared preamble}{You are part of a cross-learning system in which different users send tasks to different agents. A privacy splitter divides each graded trajectory into a reusable and a protected stream, a memory writer turns that evidence into scoped entries, and a store reviewer periodically re-scopes and merges entries. Each later run reads only the entries its user and agent may read.}

\promptbox{Privacy splitter}{%
\pr{Role.} Read one graded trajectory and decide who each part may be shared with, not how much to share. Put every detail worth keeping into exactly one of two streams.\par
\pr{Reusable.} The transferable technical lesson, phrased so that a different user on a similar task could act on it, namely what the task required, what failed, what fixed it, and what the verifier said. Keep commands, flags, error messages, and versions. Generalize an identifying detail rather than delete it, so \path{/home/alice/proj/run.sh} becomes \texttt{\textless path\textgreater/run.sh}.\par
\pr{Protected.} What is true of this user rather than of the task, such as who they are, their employer, team, project, and timezone, and their conventions for output, naming, and tools. Write it to be useful, since only this user reads it.\par
\pr{Never.} Identity numbers or the value of any credential, in either stream.\par
\pr{Output.} A reusable section and a protected section under fixed markers, with both markers always present.}

\promptbox{Memory writer}{%
\pr{Role.} Given the evidence from graded runs and the entries their user and agent can already see, decide what is worth remembering and who may read it.\par
\pr{Entry.} A \emph{situation}, saying when it applies in words a different run would recognize, and a \emph{lesson}, saying what to do with the exact filename, flag, or error text. Write only what the run taught; an empty list is a valid answer.\par
\pr{Scopes.} \GLOBAL{} is read by everyone and holds anything not tied to one user or agent; \USER{} by this user on every agent; \AGENT{} by every user on this agent; \PAIR{} by this user on this agent. Place by scope, not breadth, so a fact about a single task belongs in \GLOBAL{}. The protected call may write only \USER{} and \PAIR{}.\par
\pr{Evidence.} The recorded verdict overrides the agent's own claims; a graded failure is strong evidence, while a timeout or error establishes nothing. The benchmark's summary document and the memory files are not evidence about the task.\par
\pr{Conflicts.} When the same user wants different things from different agents, add a \PAIR{} entry and keep the other. Revise an entry the run sharpens, and delete one only when the run disproves it, with a reason.\par
\pr{Never.} Identity numbers or the value of any credential, at any scope.\par
\pr{Output.} A JSON array of entries (sample, situation, lesson, scope, entry revised) or deletions (sample, entry, reason).}

\promptbox{Store review, stage 1 (re-scoping)}{%
\pr{Input.} Every entry with its id, scope, user, agent, stream, situation, and lesson, but no counts or outcomes, so the text alone is the evidence.\par
\pr{Moves.} One step at a time, widening from \PAIR{} to \USER{} or \AGENT{} and from either to \GLOBAL{}, or narrowing back. Ask whom the lesson serves rather than who produced it; two entries stating one fact under different scopes are a reason to widen both.\par
\pr{Narrowing.} Narrowing is not free. Keep a convention the user wants from every agent at \USER{}, and move it to \PAIR{} only when the user wants something different on another agent.\par
\pr{Privacy cap.} A protected entry may sit at \PAIR{} or \USER{}, never at \AGENT{} or \GLOBAL{}.\par
\pr{Output.} A JSON array of changed entries only (entry, current scope, new scope, a short rationale).}

\promptbox{Store review, stage 2 (merging)}{%
\pr{Input.} The entries of one destination, sharing scope and owner, in similarity-ordered slices when the destination is large.\par
\pr{Rule.} Merge only entries that describe the same situation and recommend the same behavior; similar wording is not enough. Keep apart a do and a do-not about one tool, facts that hold under different conditions, and a hedged claim and a flat one.\par
\pr{Text.} Remove redundant wording only, keep the exact specifics, and add nothing new; the system carries every origin over.\par
\pr{Output.} A JSON array of merges (entries, merged situation, merged lesson, a short rationale); an empty list is common.}

\promptbox{Privacy judge}{%
\pr{System.} You are a privacy leak auditor. Given DATA and one question about a user, decide using only the data, and reply YES or NO.\par
\pr{Question.} Does DATA reveal the candidate fact, including through a paraphrase or equivalent wording? The candidate is a comparison value, not evidence.}

\section{Use of AI Assistants}
\label{app:ai-use}
We used Claude Code, an AI coding assistant, to implement, test, and debug the benchmark, the memory system, and the analysis and plotting code, and to edit drafts of this paper for clarity and consistency.
The authors designed the study and its experiments, reviewed all assisted code and text, and take responsibility for every claim and number reported.
This statement does not cover the LLMs that are objects of study, the agents and the memory components, which Sec.~\ref{sec:experiment-setup} and Appendix~\ref{app:situated} describe.

\end{document}